\documentclass{article}
\usepackage{ijcai25}

\usepackage{times}
\usepackage{soul}
\usepackage{url}
\usepackage[hidelinks]{hyperref}
\usepackage[utf8]{inputenc}
\usepackage[small]{caption}
\usepackage{graphicx}
\usepackage{amsthm}
\usepackage{amsmath}
\usepackage{amssymb}
\usepackage{booktabs}
\usepackage[switch]{lineno}
\usepackage[linesnumbered,ruled,vlined]{algorithm2e}

\theoremstyle{plain}
\newtheorem{theorem}{Theorem}[section]

\newtheorem{lemma}[theorem]{Lemma}

\theoremstyle{definition}
\newtheorem{definition}[theorem]{Definition}

\theoremstyle{remark}

\usepackage{graphicx}    
\usepackage{caption}     
\usepackage{subcaption}  
\usepackage{array}       
\usepackage{color, xcolor}
\usepackage{stackengine}
\usepackage{multirow} 
\usepackage{tabularray}
\usepackage{hyperref}
\usepackage{pifont}

\title{Think Twice before Adaptation: Improving Adaptability of DeepFake Detection via Online Test-Time Adaptation}

\author{
Hong-Hanh Nguyen-Le$^1$
\and
Van-Tuan Tran$^2$\and
Dinh-Thuc Nguyen$^{3}$\And
Nhien-An Le-Khac$^1$\\
\affiliations
$^1$ University College Dublin, Ireland\\
$^2$ Trinity College Dublin, Ireland\\
$^3$ University of Science, Ho Chi Minh City, Vietnam\\
\emails
hong-hanh.nguyen-le@ucdconnect.ie,
tranva@tcd.ie,
ndthuc@fit.hcmus.edu.vn,
an.lekhac@ucd.ie 
}

\begin{document}

\maketitle

\begin{abstract}
    Deepfake (DF) detectors face significant challenges when deployed in real-world environments, particularly when encountering test samples deviated from training data through either postprocessing manipulations or distribution shifts. We demonstrate postprocessing techniques can completely obscure generation artifacts presented in DF samples, leading to performance degradation of DF detectors. To address these challenges, we propose Think Twice before Adaptation (\texttt{T$^2$A}), a novel online test-time adaptation method that enhances the adaptability of detectors during inference without requiring access to source training data or labels. Our key idea is to enable the model to explore alternative options through an Uncertainty-aware Negative Learning objective rather than solely relying on its initial predictions as commonly seen in entropy minimization (EM)-based approaches. We also introduce an Uncertain Sample Prioritization strategy and Gradients Masking technique to improve the adaptation by focusing on important samples and model parameters. Our theoretical analysis demonstrates that the proposed negative learning objective exhibits complementary behavior to EM, facilitating better adaptation capability. Empirically, our method achieves state-of-the-art results compared to existing test-time adaptation (TTA) approaches and significantly enhances the resilience and generalization of DF detectors during inference.
\end{abstract}

\section{Introduction}
Recently, Generative Artificial Intelligence (GenAI) has been used to generate DFs for malicious purposes, such as impersonation\footnote{\href{https://edition.cnn.com/2024/02/04/asia/deepfake-cfo-scam-hong-kong-intl-hnk/index.html}{Finance worker pays out $\$$25 million after video call with deepfake `chief financial officer'}} and disinformation spread\footnote{\href{https://arstechnica.com/tech-policy/2023/03/fake-ai-generated-images-imagining-donald-trumps-arrest-circulate-on-twitter/}{AI-faked images of Donald Trump’s imagined arrest swirl on Twitter}}, raising concerns about privacy and security. Several DF detection approaches have been proposed to mitigate these negative impacts \cite{nguyen2024deepfake}. Despite advances, deploying these systems in real-world environments presents two critical challenges. First, in practice, adversaries can strategically apply previously unknown postprocessing techniques to DF samples \textbf{at inference time}, completely obscuring the generation artifacts \cite{corvi2023intriguing} and successfully bypassing detection systems. Second, real-world applications are frequently exposed to test samples drawn from distributions that deviate substantially from the training data distribution \cite{pan2023dfil}, leading to performance degradation. 
To mitigate these challenges, existing approaches require access to source training data and labels for complete re-training \cite{ni2022core,shiohara2022detecting}, continual learning \cite{pan2023dfil} or test-time training \cite{chen2022ost}, which is costly and time-consuming.

In this work, we address these limitations by introducing a novel TTA-based method, namely \textbf{Think Twice before Adaptation} (\texttt{T$^2$A}), which enhances pre-trained DF detectors without requiring access to source training data or labels. Our approach achieves two key objectives: (1) enhanced resilience through dynamic adaptation to unknown postprocessing techniques; and (2) improved generalization to new samples from unknown distributions. While current TTA approaches commonly employ Entropy Minimization (EM) as the adaptation objective, solely relying on EM can result in confirmation bias caused by overconfident predictions \cite{zhang2024come} and model collapse \cite{niu2023towards}.
To this end, in T$^2$A, we design a novel Uncertainty-aware Negative Learning adaptation objective with noisy pseudo-labels, allowing the model to explore alternative options (i.e., other classes in the classification problem) rather than becoming overly confident in potentially incorrect predictions. 
%
%
For better adaptation, we incorporate Focal Loss \cite{ross2017focal} into the negative learning (NL) objective to dynamically prioritize crucial samples and propose a gradients masking technique that updates crucial model parameters whose gradients align with those of BatchNorm layers.

\textbf{Our contributions.} To the best of our knowledge, we are the first to present a novel TTA-based method for DF detection. Our contributions include:
\begin{itemize}
    \item We provide a theoretical and quantitative analysis (Sec. \ref{sec:analysis}) that demonstrates the impacts of postprocessing techniques on the detectability of DF detectors.
    \item We introduce \texttt{T$^2$A}, a novel TTA-based method specifically designed for DF detection. \texttt{T$^2$A} enables models to explore alternative options rather than relying on their initial predictions for adaptation (Sec. \ref{ssec:negative-obj}). We also theoretically demonstrate that our proposed negative learning objective exhibits complementary behavior to EM. 
    Additionally, we introduce Uncertain Sample Prioritization strategy (Sec. \ref{ssec:sample-identify}) and Gradients Masking technique (Sec. \ref{ssec:gradient-masking}) to dynamically focus on crucial samples and crucial model parameters when adapting.
    \item We evaluate \texttt{T$^2$A} under two scenarios: (i) Unknown postprocessing techniques; and (ii) Unknown data distribution and postprocessing techniques. Our experimental results show superior adaptation capabilities compared to existing TTA approaches. Furthermore, we demonstrate that integration of \texttt{T$^2$A} significantly enhances the resilience and generalization of DF detectors during inference, establishing its practical utility in real-world deployments.
\end{itemize}

\section{Related Work}\label{sec:related-work}
\subsection{Deepfake Detection}
DF detection approaches are often formulated as a binary classification problem that automatically learns discriminative features from large-scale datasets \cite{nguyen2024passive}. Existing approaches can be classified into three categories based on their inputs: (i) Spatial-based approaches that operate directly on pixel-level features \cite{ni2022core,cao2022end}, (ii) Frequency-based approaches that analyze generation artifacts in the frequency domain \cite{liu2021spatial,frank2020leveraging}, and (iii) Hybrid approaches that integrate both pixel and frequency domain information within a unified method \cite{liu2023adaptive}. Recent advances have improved the cross-dataset generalization of DF detectors by employing data augmentation (DA) strategies \cite{ni2022core,yan2024transcending}, synthesis techniques \cite{shiohara2022detecting}, continual learning \cite{pan2023dfil}, meta-learning and one-shot test-time training \cite{chen2022ost}.


Compared to existing methods, our \texttt{T$^2$A} offers advantages: (1) \texttt{T$^2$A} enables DF detectors to be adapted to test data without access to source data (e.g., OST \cite{chen2022ost} requires source data for adaptation); (2) \texttt{T$^2$A} does not rely on any DA or synthesis techniques to extend the diversity of data; (3) Not only enhance the generalization, \texttt{T$^2$A} also improves the resilience of DF detectors to unknown postprocessing techniques. Additionally, our method is orthogonal to these works \cite{fang2024unified,liu2024cfpl,he2024joint}, which require pre-training on joint datasets (physical and digital attacks) and do not adapt during inference. 

\subsection{Test-time Adaptation (TTA)}\label{ssec:related-TTA}
TTA approaches only require access to the pre-trained model from the source domain for adaptation \cite{liang2024comprehensive}. Unlike source-free domain adaptation approaches \cite{li2024comprehensive}, which require access to the entire target dataset, TTA enables online adaptation to the arrived test samples. 

TENT \cite{wang2020tent} and MEMO \cite{zhang2022memo} optimized batch normalization (BN) statistics from the test batch through EM. 
LAME \cite{boudiaf2022parameter} adapted only the model's output probabilities by minimizing Kullback–Leibler divergence between the model's predictions and optimal nearby points' vectors. Several methods have studied TTA in continuously changing environments. CoTTA \cite{wang2022continual} implemented weight and augmentation averaging to mitigate error accumulation, while EATA \cite{niu2022efficient} developed an efficient entropy-based sample selection strategy for model updates. Inspired by parameter-efficient fine-tuning, VIDA \cite{liuvida2023} used high-rank adapters to handle domain shifts. However, these methods solely rely on EM as the learning principle, which can present two issues: (1) \textbf{Confirmation bias}: EM greedily pushes for confident predictions on all samples, even when predictions are incorrect \cite{zhang2024come}, leading to overconfident yet incorrect predictions; and (2) \textbf{Model Collapse}: EM tends to cause model collapse, where the model predicts all samples to the same class, regardless of their true labels \cite{niu2023towards}. The model collapse phenomenon is particularly problematic in DF detection, where the inherent bias toward dominant fake samples in training data \cite{layton2024sok} can exacerbate the collapse.  

Focusing on the problem of EM, our \texttt{T$^2$A} method allows the model to consider alternative options rather than completely relying on its initial prediction during inference through NL with noisy pseudo-labels.  


\subsection{Negative Learning}
Supervised learning or positive learning (PL) directly maps inputs to their corresponding labels. However, when labels are noisy, PL can lead models to learn incorrect patterns. Negative learning (NL)  \cite{kim2019nlnl} addresses this challenge by training networks to identify which classes an input does not belong to. Several loss functions have been proposed by leveraging this concept: NLNL \cite{kim2019nlnl} combines sequential PL and NL phases, while JNPL \cite{kim2021joint} proposes a single-phase approach through joint optimization of enhanced NL and PL loss functions. Recent work has further integrated NL principles with normalization techniques \cite{ma2020normalized} to transform active losses into passive ones \cite{ye2023active}.

Inspired by these advances, we introduce a NL strategy with noisy pseudo-labels to our \texttt{T$^2$A} method to enable the model to think twice during adaptation, avoiding confirmation bias and model collapse caused by EM.

\section{Generation Artifacts Analysis}\label{sec:analysis}
Artifacts in DFs generated by Generative Adversarial Examples (GANs), which emerge from the upsampling operations in the GANs pipeline, can be revealed in the frequency domain through Discrete Fourier Transform (DFT) \cite{frank2020leveraging}. In this section, we demonstrate postprocessing techniques can completely obscure these artifacts presented in DF samples, leading to performance degradation of DF detectors.


\begin{definition}
\label{def:dct}
Let an image $x(\cdot,\cdot)$ of size $M \times N$, its DFT $X(\cdot,\cdot)$ is defined as:

\begin{equation}
    X(u, v) = \frac{1}{MN} \sum^{M-1}_{m=0} \sum^{N-1}_{n=0} x(m,n)e^{-j2\pi(\frac{um}{M}+ \frac{vn}{N})},
\end{equation}
where $x(m,n)$ represents pixel values at spatial coordinates and $X(u,v)$ denotes the corresponding Fourier coefficient in frequency domain.
\end{definition}

\begin{lemma}
    \label{lemma:conv-frequency}
    For two images $x_1(\cdot,\cdot)$ and $x_2(\cdot,\cdot)$, their convolution in the spatial domain is equivalent to multiplication of their spectra in the frequency domain:
    \begin{equation}
        x_1(m,n) \circledast x_2(m,n) \Leftrightarrow X_1(u, v) \cdot X_2(u,v).
    \end{equation}
\end{lemma}
This property (Proof in Appendix \ref{sec:proofs}) is particularly important for understanding why the upsampling operation leaves artifacts in the frequency domain \cite{ojha2023towards}. For an image $x(\cdot,\cdot)$ convolved with a kernel $c(\cdot,\cdot)$, the output $y(\cdot,\cdot)$ in the spatial domain and its frequency domain form can be expressed as:
\begin{equation}
    \begin{split}
    y(m,n) &= x(m,n) \circledast c(m,n) \\
    \Leftrightarrow Y(u,v) &= X(u,v) \cdot C(u,v)
    \end{split}
\end{equation}

\begin{figure}[!ht]
    \centering
    \begin{tabular}{ccccc}
        Real & Fake & Resize & Gaussian Blur \\
        \includegraphics[width=0.1\textwidth]{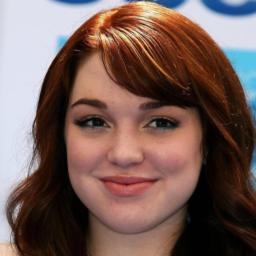} &
        \includegraphics[width=0.1\textwidth]{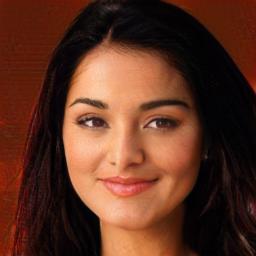} &
        \includegraphics[width=0.1\textwidth]{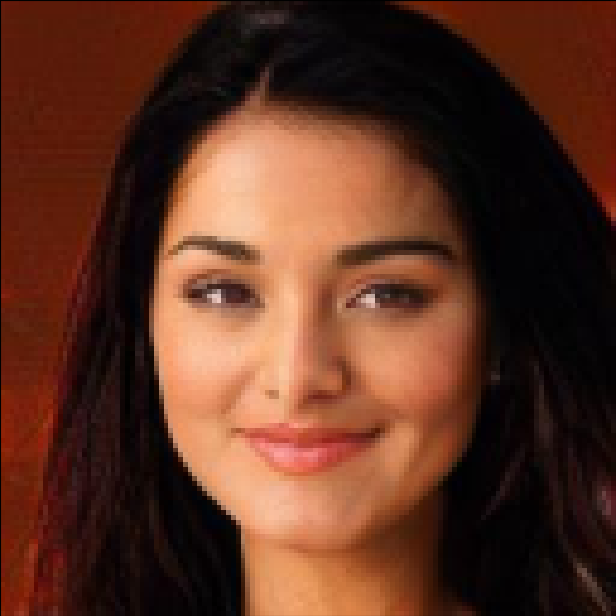} &
        \includegraphics[width=0.1\textwidth]{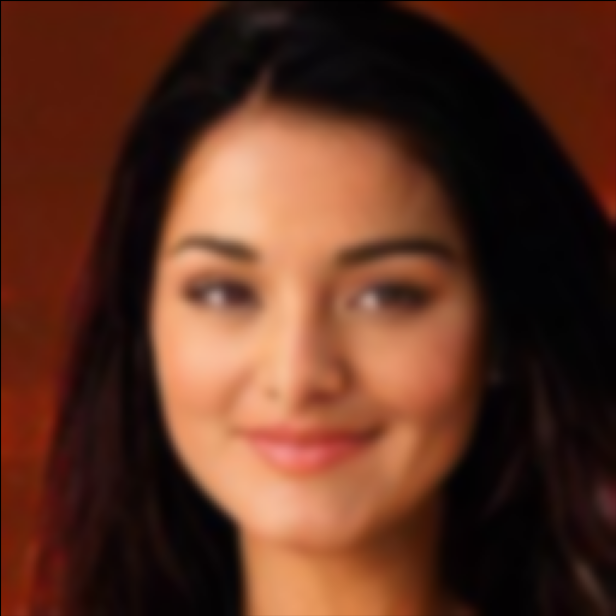} \\
        \includegraphics[width=0.1\textwidth]{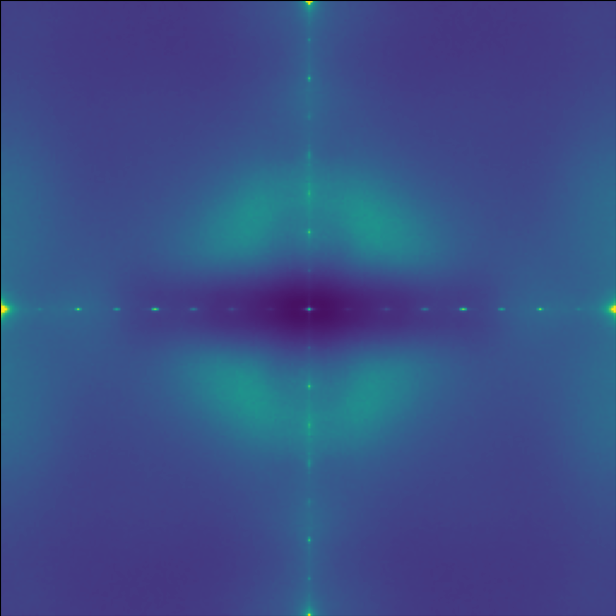} &
        \includegraphics[width=0.1\textwidth]{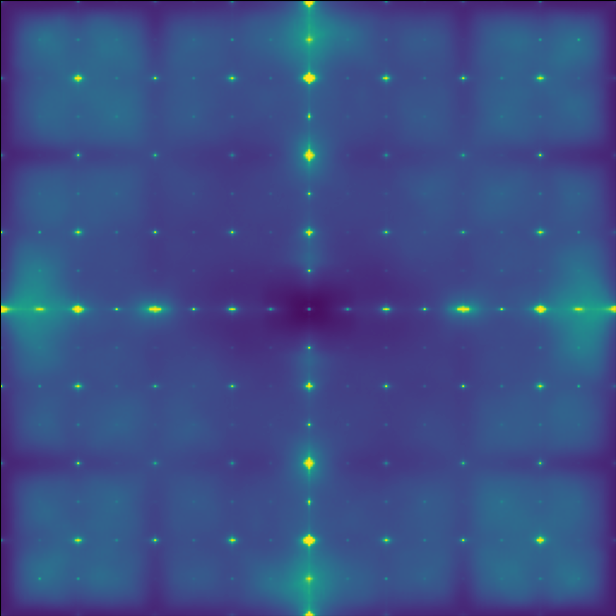} &
        \includegraphics[width=0.1\textwidth]{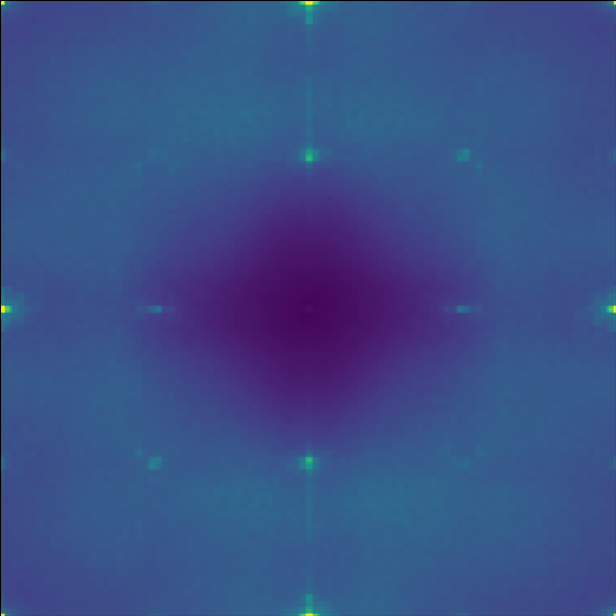} &
        \includegraphics[width=0.1\textwidth]{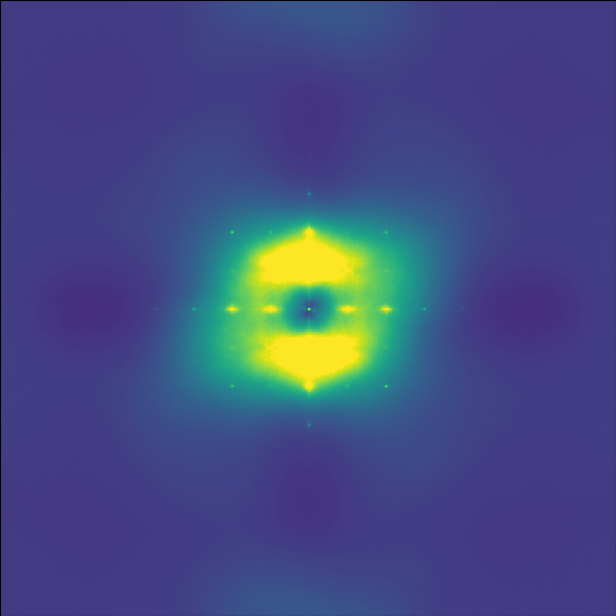} \\
        (a) & (b) & (c) & (d) 
    \end{tabular}
    \caption{Comparison of frequency domain artifacts across different image processing conditions. Top row: Images in spatial domain. Bottom row: Corresponding frequency spectra. Artifacts as \textit{checkerboard patterns} in (c) and (d) are obscured by postprocessing techniques (i.e., Resize, Gaussian Blur). All fake images are generated by StarGANv2.}
    \label{fig:spectral_comparison}
\end{figure}

When image $x(\cdot,\cdot)$ is upsampled by a factor of $2$ in both dimensions, the upsampled image $\tilde{x}(\cdot,\cdot)$ can be expressed as:

\begin{equation}
    \tilde{x}(m,n) =
  \begin{cases}
    x\left(\frac{m}{2}, \frac{n}{2}\right),       & \quad m = 2k, n = 2l\\
    0  & \quad \text{otherwise}.
  \end{cases}
\end{equation}
where $k=0, \dots, M-1$ and $l=0,\dots,N-1$. The DFT of the upsampled image is:
\begin{equation}
    \tilde{X}(u, v) = \frac{1}{4MN} \sum^{2M-1}_{m=0} \sum^{2N-1}_{n=0} \tilde{x}(m,n)e^{-j2\pi(\frac{um}{2M}+ \frac{vn}{2N})}
\end{equation}
This upsampling operation creates a characteristic periodic structure in the frequency domain, showing that the original image's frequency components appear multiple times in the frequency domain:
\begin{equation}
    \resizebox{.98\linewidth}{!}{$
            \displaystyle
    \tilde{X}(u, v) =
  \begin{cases}
    X(u,v),       & \; u \in [0, M-1], v \in [0, N-1] \\
    X(u-M,v),  & \; u \in [M, 2M-1], v \in [0, N-1] \\
    X(u, v-N), & \; u \in [0, M-1], v \in [N, 2N-1] \\
    X(u-N, v-N), & \;  u \in [M, 2M-1], v \in [N, 2N-1]
  \end{cases}
  $}
\end{equation}
These duplicated components create distinctive artifacts as \textit{checkerboard patterns} in the frequency domain that distinguishes GAN-generated images from real ones. 

However, these spectral artifacts exhibit vulnerability to various postprocessing operations \cite{corvi2023intriguing}. As shown in Figure \ref{fig:spectral_comparison}(b), the GAN-generated image displays distinctive checkerboard artifacts in its frequency spectrum, but they undergo substantial modifications when subjected to different postprocessing operations (Figures \ref{fig:spectral_comparison}(c)-(d)). The magnitude of these artifacts' obscurity correlates directly with the intensity of the applied postprocessing operations, as demonstrated in Figure \ref{fig:more-spectral-comparison} (Appendix \ref{sec:more-artifacts}). Furthermore, the empirical analysis presented in Figure \ref{fig:exp_cross_augs} of Appendix \ref{sec:more-artifacts} shows that the performance of existing DF detectors tends to drop significantly when encountering unseen postprocessing techniques with increasing intensities. 

\section{Methodology}\label{sec:method}

The core principle of \texttt{T$^2$A} lies in its deliberate approach to decision-making, encouraging models to explore alternative options rather than solely relying on their initial predictions. 
The key steps of \texttt{T$^2$A} are summarized in Algorithm \ref{alg:T2A}.

\subsection{Problem Definition}\label{ssec:problem-def}
Given a DF detector $f: \mathcal{X} \rightarrow \mathbb{R}^2$ parameterized by $\theta$ is well-trained on the training data $\mathcal{D}^{train} = \{(x_i, y_i)\}^{N^{train}}_{i=1} \sim P^{train}(x,y)$, where $x \in \mathcal{X}$ is the input and $y \in \mathcal{Y}=\{0,1\}$ is the target label, our goal is to online update parameters $\theta$ of $f$ on mini-batches $\{\mathcal{B}_1, \mathcal{B}_2, \dots\}$ of the test stream $\mathcal{D}^{test} = \{(x_j, y_j)\}^{N^{test}}_{j=1} \sim P^{test}(x,y)$. Note that, in the online TTA setting, $P^{train}(x,y)$ and $\{y_j\}$ are unavailable, and the knowledge learned in previously seen mini-batches could be accumulated for adaptation to the current mini-batch \cite{liang2024comprehensive}.  In this work, we consider online TTA in two challenging scenarios of DF detection:
\begin{enumerate}
    \item \textbf{Unseen postprocessing Techniques}: While the test data distribution remains similar to the training distribution $P^{train}(x,y) = P^{test}(x,y)$, the test samples are applied unknown postprocessing operations $\Psi: \mathcal{X} \rightarrow \mathcal{X}$. Specifically, given a test sample $x_j \sim P^{test}$, $f$ takes $\Psi(x_j)$ as input, where $\Psi \in \mathfrak{P}$ with $\mathfrak{P}$ denotes a set of unseen postprocessing techniques during training.
    \item \textbf{Unseen Data Distribution and postprocessing Techniques}: This is a more challenging setting in which test samples come from a different distribution $P^{test} \neq P^{train}$ and are also subjected to unknown postprocessing operations.
\end{enumerate}



\SetKwComment{Comment}{\#}{}
\begin{algorithm}[ht]
    \small
    \DontPrintSemicolon
    \SetKwInOut{Input}{Input}
    \SetKwInOut{Define}{Define}
    \SetKwFunction{LocalTraining}{LocalTraining}
    \SetKwFunction{SelfPacedTransform}{SelfPacedTransform}
    \SetKwProg{Fn}{Function}{:}{}
    
    \Input{trained model $f_\theta$, test samples $\mathcal{D}^{test} = \{x_j, y_j)\}^{N^{test}}_{j=1}$}
    \Define{batch size $B$; loss balancing hyperparameters $\alpha, \beta$, gradients alignment threshold $\psi$; learning rate $\eta$}
    \For{mini-batches $\{x_i\}_{i=1}^B \subset \mathcal{D}^{test}$}{
        Obtain pseudo-label $\hat{y_i}$ from Eq. \ref{eq:pseudo-label} \;
        Calculate noisy pseudo-label by Eq. \ref{eq:noisy_pseudo_labels}\; 
        Calculate entropy of model predictions $\mathcal{L}_{EM}$ follow Eq. \ref{eq:EM} \;
        Calculate noise-tolerant negative loss $\mathcal{L}_{NTNL}(x_i, \tilde{y}_i) = \alpha \mathcal{L}_{nn}(x_i, \tilde{y}_i) + \beta \mathcal{L}_{p}(x_i, \tilde{y}_i)$ follow Equations (\ref{eq:nn}) and (\ref{eq:NTNL}) \;
        Optimize the adaptation objective function: $\mathcal{L}_{NTNL} + \mathcal{L}_{EM}$ to obtain the gradient matrix $\nabla_\theta\mathcal{L}$ \;
        Perform Gradient Masking on $\nabla_\theta\mathcal{L}$ by keeping the parameters of those gradients aligned with gradients of BN layers by Eq. \ref{eq:gradients_filtering} \;
        Perform Gradient Descent to adapt the model: $\theta \gets \theta - \eta \nabla_\theta\mathcal{L}$ \;
    }    

    \caption{T$^2$A Algorithm}
    \label{alg:T2A}
\end{algorithm}

\subsection{Revisitting Entropy Minimization (EM)} \label{ssec:EM}
EM is commonly used to update model parameters by minimizing the entropy of model outputs on test sample $x$ during inference:
\begin{equation}
    \mathcal{L}_{EM} = - \sum_{c \in C}p(y=c|x)\log p(y=c|x),
    \label{eq:EM}
\end{equation}
where $p(y=c|x)$ is the predicted probability for class $c$, computed as the softmax output of the model: $p(y=c|x)= \frac{\exp(f_c(x))}{\sum_{c \in C} \exp(f_j(x)}$, where $f_c(x)$ is the logit for class $c$ from the model's forward pass on input $x$. As discussed in Sec \ref{ssec:related-TTA}, EM causes two issues: Confirmation bias and Model collapse. Therefore, besides EM, our \texttt{T$^2$A} method introduces a NL strategy with noisy pseudo-labels (described in Sec. \ref{ssec:negative-obj}), allowing models to re-think other potential options before making the final decision. 


%

\subsection{Uncertainty-aware Negative Learning}\label{ssec:negative-obj}
\subsubsection{Uncertainty Modelling with Noisy Pseudo-Labels}\label{subsec:noisy-labels}

Given the DF detector $f$, the pseudo-label  $\hat{y} = \hat{y}(x) \in \{0,1\}$ of input $x$ is defined as:
\begin{equation}
\hat{y} = \text{sign}(f(x) - \tau) = 
    \begin{cases}
        1, \quad f(x) \geq \tau \\
        0, \quad f(x) < \tau
    \end{cases},
    \label{eq:pseudo-label}
\end{equation}
where $\tau \in [0, 1]$ denotes the classification threshold. Rather than implicitly trusting the model's initial predictions, we enable the model to "doubt" its predictions by introducing noisy pseudo-labels.

We model the uncertainty in pseudo-labels using a Bernoulli distribution. For each input $x$ with pseudo-label $\hat{y}$, we generate a noisy pseudo-label $\tilde{y}$ for input $x_i$ as follows:

\begin{equation}
\tilde{y} = \begin{cases}
1 - \hat{y}, & \text{if } X \sim \texttt{Bernoulli}(1-p_{x_i}) = 1 \\
\hat{y}, & \text{otherwise}
\end{cases},
\label{eq:noisy_pseudo_labels}
\end{equation}
where $p_{x_i}$ represents the prediction probability. This indicates that higher confidence predictions have a lower probability of being flipped. When the Bernoulli trial equals $1$ (with probability $1-p_{x_i}$), the pseudo-label is flipped to the opposite class; otherwise (with probability $p_{x_i}$), it remains unchanged. However, directly adapting to noisy pseudo-labels presents two limitations during test-time updates: (1) Without access to source data for regularization, errors from noisy labels can accumulate rapidly; and (2) The stochastic nature of noisy gradients can lead to unstable updates.



\subsubsection{Noise-tolerant Negative Loss Function}\label{ssec:NTNL}

The goal of the noise-tolerant negative loss (NTNL) is to enable the model to think twice through NL with noisy pseudo-labels. 

\textbf{From Positive to Negative Learning.} Negative learning (NL) enables the model to be taught with a lesson that $"$\textit{this input image does not belong to this complementary label"} \cite{kim2019nlnl}. In our work, converting from pseudo-labels to noisy versions is equivalent to transforming from positive to negative learning, facilitating the DF model to re-think that "this input image might not belong to this real (fake)/fake (real) label$"$. 

\textbf{Noise-tolerant Negative Loss Function.} Inspired by existing works \cite{zhou2021asymmetric,ma2020normalized,ghosh2017robust}, we start from the fact that any loss function can be robust to noisy labels through a simple normalization operation:
\begin{equation}
    \mathcal{L}_{norm} = \frac{\ell(f(x),y)}{\sum_{c \in C}\ell(f(x), c)}.
    \label{eq:norm}
\end{equation}

\begin{theorem}
    In the binary classification with pseudo-label $\hat{y} \in \{0,1\}$, if the normalized loss function $\mathcal{L}_{norm}$ has the local extremum at $x^*$, the entropy minimization function $\mathcal{L}_{EM}$ also has the local at $x^{*}$, and vice versa.
    \label{theorem:norm-entropy}
\end{theorem}

From Theorem \ref{theorem:norm-entropy} (Proof in Appendix \ref{sec:proofs}), we demonstrate that simply using pseudo-labels in the normalized loss function could drive the model toward maximizing confidence in its initial predictions $\hat{y}$. This behavior aligns with the EM objective presented in Eq.\ref{eq:EM}. 
However, we seek to enable the model to explore another option rather than uncritically trusting its initial predictions, which may be incorrect. 
To do that, we introduce noisy pseudo-labels $\tilde{y}$ in place of the original pseudo-labels $\hat{y}$ within the normalized loss function, in which $\tilde{y}$ is generated by the flipping procedure described previously, effectively transforming normalized loss function (Eq. \ref{eq:norm}) to a negative one. This normalized negative loss $\mathcal{L}_{nn}$ for adapting with noisy pseudo-labels is defined as:



\begin{equation}
    \mathcal{L}_{nn}(x,\tilde{y}) = \frac{\ell(f(x),\tilde{y})}{\sum_{c=\in \{0,1\}}{\ell(f(x), c)}}.
    \label{eq:nn}
\end{equation}

As shown in Figure \ref{fig:proposed-loss} (Appendix \ref{sssection:analysis-losses}), given a normalized loss function with pseudo label $\mathcal{L}_{norm}(x, \hat{y})$, our normalized negative loss function $\mathcal{L}_{nn}(x,\tilde{y})$ with noisy pseudo-label is the opposite of $\mathcal{L}_{norm}(x, \hat{y})$. 

Prior research by \cite{ma2020normalized,ye2023active} has indicated that the normalized loss function suffers from the underfitting problem.
This problem is particularly critical in the TTA context where the model only "sees" a few samples during inference.
To address this challenge, we incorporate the passive loss function $\mathcal{L}_{p}$ \cite{ye2023active} into TTA, leading to our NTNL which can effectively help the model to adapt to noisy pseudo-labels:


\begin{equation}
    \mathcal{L}_{NTNL}(x, \tilde{y}) = \alpha  \mathcal{L}_{nn}(x, \tilde{y}) +  \beta\mathcal{L}_{p}(x, \tilde{y}),
\label{eq:NTNL}
\end{equation}
where $\mathcal{L}_{p}(x, \tilde{y}) = 1 - \frac{p_0 - \ell(f(x),\tilde{y})}{\sum_{c \in \{0,1\}}{p_0 - \ell(f(x), c)}},$
$p_0$ is the minimum value of the model prediction in the current test batch, and $\alpha, \beta$ are balancing hyperparameters. 
%
\begin{definition} (Passive loss function).
    $\mathcal{L}_{p}$ is a passive loss function if $\forall(x,y) \in \mathcal{D}, \exists k \neq y, \ell(f(x), k) \neq 0.$
\end{definition}

\subsection{Uncertain Sample Prioritization} \label{ssec:sample-identify}

To identify which samples should be prioritized during adaptation,  we propose a dynamic prioritization strategy that focuses on uncertain samples (i.e., low confidence). Our intuition here is that lower-confidence samples require the model to be considered more carefully. Specifically, we incorporate Focal Loss \cite{ross2017focal} into the NTNL function (Eq. \ref{eq:NTNL}). Formally, the loss function $\ell(x, \tilde{y})$ is now defined: 
\begin{equation}
    \ell(x, \tilde{y}) = -(1-p(\tilde{y}|x)^{\gamma})\log p(\tilde{y} |x),
\label{eq:focal_loss}
\end{equation}
where $\gamma$ controls the rate at which high-confident samples are down-weighted.

The proposed NTNL with Focal Loss enables the model to explore alternative options beyond its initial predictions while dynamically focusing on uncertain samples during adaptation. When combined with EM, we formulate our final adaptation objective function to enhance the adaptation of DF detectors as follows:

\begin{equation}
    \mathcal{L} = \mathcal{L}_{NTNL} + \mathcal{L}_{EM},
\end{equation}
where $\mathcal{L}_{EM}$ is the entropy of model predictions defined in Eq. \ref{eq:EM}. By optimizing this objective, our approach achieves robust adaptation that can effectively handle both unknown postprocessing techniques and distribution shifts during inference.

\subsection{Gradients Masking}\label{ssec:gradient-masking}
BatchNorm (BN) adaptation \cite{schneider2020improving} is widely used in existing TTA approaches \cite{niu2022efficient,wang2020tent}. BN is a crucial layer that normalizes each feature $z$ during training: $y = \varrho * \left(\frac{(z - \mu^b)}{\sigma^b}\right) + \vartheta$, where $\mu^b$ and $\sigma^b$ are batch statistics, and $\varrho$, $\vartheta$ are learnable parameters. After training, $\mu^{ema}$ and $\sigma^{ema}$, which are estimated over the whole training dataset via exponential moving average (EMA) \cite{schneider2020improving}, are used during inference. When $P^{train}(x,y) \neq P^{test}(x,y)$, BN adaptation replaces EMA statistics ($\mu^{ema}$, $\sigma^{ema}$) with statistics computed from test mini-batches ($\hat{\mu}^{b}$, $\hat{\sigma}^{b}$). However, this approach is limited by only updating BN layer parameters.

To overcome this limitation, we propose a gradient masking technique that identifies and updates parameters whose gradients align with those of BN layers. Let $\theta_{BN_i}$ be the parameter of $i$-th BN layer, and all BN parameters' gradients are concatenated into a single vector: $ u = [\nabla_{\theta_{BN_1}}\mathcal{L}, \nabla_{\theta_{BN_2}}\mathcal{L}, ..., \nabla_{\theta_{BN_L}}\mathcal{L}],$
where $N$ is the number of BN layers and $\nabla_{\theta_{BN_i}}\mathcal{L}$ represents the gradient vector of the loss $\mathcal{L}$ with respect to parameters in the $i$-th BN layer. For each non-BN parameter's gradient $v_i = \nabla_{\theta_i}\mathcal{L}$ in the model, we compute its cosine similarity with the concatenated BN gradients: $\text{sim}(u, v_i) = \frac{\langle v_i, u \rangle}{||v_i|| \cdot ||u||}$.

Note that, since parameter gradients and BN gradient vectors have different dimensions, zero-padding is applied to align dimensions before computing similarity. The final gradient masking is then applied as:

\begin{equation}
    \nabla_{\theta_i}\mathcal{L} = \begin{cases} 
v_i & \text{if } \text{sim}(v_i, u) > \psi \\
0 & \text{otherwise}
\end{cases},
\label{eq:gradients_filtering}
\end{equation}
where $\psi$ is a threshold to control the selection of parameters for updating. This technique brings more capacity for adaptation as more model parameters are updated compared to approaches that only update BN parameters during inference \cite{niu2022efficient,wang2020tent}.

\begin{table*}[ht]
\centering
\fontsize{6pt}{6pt}\selectfont
\caption{Comparison with state-of-the-art TTA methods on FF++ with different unknown postprocessing techniques. The results for each postprocessing technique are averaged across 5 intensity levels. Bold values denote the best performance for each metric.}
\label{tab:TTA-post-process}
\begin{tblr}{
  colspec = {Q[70]Q[35]Q[35]Q[35]Q[35]Q[35]Q[35]Q[35]Q[35]Q[35]Q[35]Q[35]Q[35]Q[35]Q[35]Q[35]},
  cells = {c},
  cell{1}{1} = {r=3}{},
  cell{1}{2} = {c=15}{0.15\linewidth},
  cell{2}{2} = {c=3}{0.15\linewidth},
  cell{2}{5} = {c=3}{0.15\linewidth},
  cell{2}{8} = {c=3}{0.13\linewidth},
  cell{2}{11} = {c=3}{0.13\linewidth},
  cell{2}{14} = {c=3}{0.13\linewidth},
  vline{2} = {1}{},
  vline{2,5,8,11,14} = {2-12}{},
  hline{1,13} = {-}{0.08em},
  hline{2-3} = {2-16}{},
  hline{4,12} = {-}{},
  hline{4} = {2}{-}{},
}
\textbf{ Method } & \textbf{Postprocessing Techniques} &                                     &                                     &                                     &                                     &                                     &                                     &                                     &                                     &                                     &                                     &                                     &                                     &                                     &                                     \\
                  & \textbf{Color Contrast }            &                                     &                                     & \textbf{Color Saturation}           &                                     &                                     & \textbf{Resize}                     &                                     &                                     & \textbf{Gaussian Blur}              &                                     &                                     & \textbf{Average}                    &                                     &                                     \\
                  & \textbf{ACC}                        & \textbf{AUC}                        & \textbf{AP}                         & \textbf{ACC}                        & \textbf{AUC}                        & \textbf{AP}                         & \textbf{ACC}                        & \textbf{AUC}                        & \textbf{AP}                         & \textbf{ACC}                        & \textbf{AUC}                        & \textbf{AP}                         & \textbf{ACC}                        & \textbf{AUC}                        & \textbf{AP}                         \\
Source            & 0.7891 $\pm$ 0.04                   & 0.8696 $\pm$ 0.03                   & 0.9639 $\pm$ 0.01                   & 0.8074 $\pm$ 0.04                   & 0.8195 $\pm$ 0.06                   & 0.9432 $\pm$ 0.02                   & 0.8120 $\pm$ 0.03                   & 0.8767 $\pm$ 0.02                   & 0.9669 $\pm$ 0.01                   & 0.8431 $\pm$ 0.01                   & 0.8423 $\pm$ 0.04                   & 0.9523 $\pm$ 0.01                   & 0.8129 $\pm$ 0.01                   & 0.8520 $\pm$ 0.02                   & 0.9566 $\pm$ 0.01                   \\
TENT              & 0.8745 $\pm$ 0.01                   & 0.9043 $\pm$ 0.01                   & 0.9732 $\pm$ 0.01                   & 0.8408 $\pm$ 0.03                   & 0.8510 $\pm$ 0.05                   & 0.9562 $\pm$ 0.01                   & \textbf{0.8517} $\pm$ 0.01 & 0.8837 $\pm$ 0.02                   & 0.9680 $\pm$ 0.01                   & 0.8622 $\pm$ 0.01                   & 0.8844 $\pm$ 0.02                   & 0.9676 $\pm$ 0.01                   & 0.8573 $\pm$ 0.01                   & 0.8808 $\pm$ 0.01                   & 0.9663 $\pm$ 0.01                   \\
MEMO              & 0.8288 $\pm$ 0.01                   & 0.8612 $\pm$ 0.01                   & 0.9603 $\pm$ 0.01                   & 0.8268 $\pm$ 0.01                   & 0.8244 $\pm$ 0.04                   & 0.9482 $\pm$ 0.01                   & 0.8348 $\pm$ 0.01                   & 0.8611 $\pm$ 0.02                   & 0.9620 $\pm$ 0.01                   & 0.8334 $\pm$ 0.01                   & 0.8676 $\pm$ 0.02                   & 0.9626 $\pm$ 0.01                   & 0.8310 $\pm$ 0.01                   & 0.8536 $\pm$ 0.01                   & 0.9583 $\pm$ 0.01                   \\
EATA              & 0.8740 $\pm$ 0.01                   & 0.9044 $\pm$ 0.01                   & 0.9733 $\pm$ 0.01                   & 0.8402 $\pm$ 0.03                   & 0.8507 $\pm$ 0.05                   & 0.9561 $\pm$ 0.01                   & 0.8511 $\pm$ 0.01                   & 0.8839 $\pm$ 0.02                   & 0.9681 $\pm$ 0.01                   & 0.8625 $\pm$ 0.01                   & 0.8846 $\pm$ 0.02                   & 0.9676 $\pm$ 0.01                   & 0.8570 $\pm$ 0.01                   & 0.8809 $\pm$ 0.01                   & 0.9663 $\pm$ 0.01                   \\
CoTTA             & 0.8548 $\pm$ 0.01                   & 0.8706 $\pm$ 0.02                   & 0.9596 $\pm$ 0.01                   & 0.8214 $\pm$ 0.01                   & 0.8256 $\pm$ 0.01                   & 0.9481 $\pm$ 0.01                   & 0.8445 $\pm$ 0.01                   & 0.8618 $\pm$ 0.02                   & 0.9618 $\pm$ 0.01                   & 0.8517 $\pm$ 0.01                   & 0.8664 $\pm$ 0.02                   & 0.9622 $\pm$ 0.01                   & 0.8431 $\pm$ 0.01                   & 0.8561 $\pm$ 0.01                   & 0.9579 $\pm$ 0.01                   \\
LAME              & 0.7882 $\pm$ 0.03                 & 0.8185 $\pm$ 0.05                 & 0.9393 $\pm$ 0.01                 & 0.8088 $\pm$ 0.03                 & 0.7594 $\pm$ 0.05                 & 0.9096 $\pm$ 0.03                 & 0.7957 $\pm$ 0.01                 & 0.8113 $\pm$ 0.02                 & 0.9311 $\pm$ 0.01                 & 0.8065 $\pm$ 0.01                 & 0.7519 $\pm$ 0.06                 & 0.9035 $\pm$ 0.02                 & 0.7998 $\pm$ 0.01                 & 0.7853 $\pm$ 0.02                 & 0.9209 $\pm$ 0.01                 \\
VIDA              & 0.8517 $\pm$ 0.01                   & 0.8794 $\pm$ 0.01                   & 0.9647 $\pm$ 0.01                   & 0.8168 $\pm$ 0.02                   & 0.8210 $\pm$ 0.05                   & 0.9446 $\pm$ 0.01                   & 0.8385 $\pm$ 0.01                   & 0.8668 $\pm$ 0.03                   & 0.9617 $\pm$ 0.01                   & 0.8448 $\pm$ 0.01                   & 0.8631 $\pm$ 0.02                   & 0.9596 $\pm$ 0.01                   & 0.8380 $\pm$ 0.01                   & 0.8576 $\pm$ 0.01                   & 0.9576 $\pm$ 0.01                   \\
COME              & 0.8660 $\pm$ 0.01                 & 0.8983 $\pm$ 0.01                 & 0.9716 $\pm$ 0.01                 & 0.8391 $\pm$ 0.02                 & 0.8502 $\pm$ 0.05                 & 0.9568 $\pm$ 0.02                 & 0.8528 $\pm$ 0.02                 & 0.8781 $\pm$ 0.03                 & 0.9654 $\pm$ 0.01                 & 0.8622 $\pm$ 0.01                 & 0.8812 $\pm$ 0.02                 & 0.9665 $\pm$ 0.01                 & 0.855 $\pm$ 0.01                  & 0.877 $\pm$ 0.02                  & 0.9651 $\pm$ 0.01                 \\
\texttt{T$^2$A} (Ours)        & \textbf{0.8745} $\pm$ 0.01 & \textbf{0.9044} $\pm$ 0.02 & \textbf{0.9733} $\pm$ 0.01 & \textbf{0.8437} $\pm$ 0.03 & \textbf{0.8519} $\pm$ 0.05 & \textbf{0.9566} $\pm$ 0.01 & 0.8502 $\pm$ 0.02                   & \textbf{0.8840} $\pm$ 0.02 & \textbf{0.9681} $\pm$ 0.01 & \textbf{0.8642} $\pm$ 0.01 & \textbf{0.8847} $\pm$ 0.02 & \textbf{0.9676} $\pm$ 0.01 & \textbf{0.8582} $\pm$ 0.01 & \textbf{0.8813} $\pm$ 0.01 & \textbf{0.9664} $\pm$ 0.01 
\end{tblr}
\end{table*}

\begin{table*}[ht]
\centering
\fontsize{6pt}{6pt}\selectfont
\caption{Comparison with state-of-the-art TTA methods under the unknown data distributions and postprocessing techniques scenario across 6 deepfake datasets. Bold values denote the best performance for each metric.}
\label{tab:TTA-cross-dataset}
\begin{tblr}{
  colspec = {Q[50]Q[35]Q[35]Q[35]Q[35]Q[35]Q[35]Q[35]Q[35]Q[35]Q[35]Q[35]Q[35]Q[35]Q[35]Q[35]Q[35]Q[35]Q[35]},
  cells = {c},
  cell{1}{1} = {r=2}{},
  cell{1}{2} = {c=3}{0.1\linewidth},
  cell{1}{5} = {c=3}{0.1\linewidth},
  cell{1}{8} = {c=3}{0.11\linewidth},
  cell{1}{11} = {c=3}{0.1\linewidth},
  cell{1}{14} = {c=3}{0.1\linewidth},
  cell{1}{17} = {c=3}{0.1\linewidth},
  vline{2} = {1}{},
  vline{2,5,8,11,14,17} = {1-11}{},
  hline{1,12} = {-}{0.08em},
  hline{2} = {2-19}{},
  hline{3,11} = {-}{},
  hline{3} = {2}{-}{},
}
\textbf{Mehtod} & \textbf{CelebDF-v1}   &                       &                       & \textbf{CelebDF-v2}   &                       &                       & \textbf{DFD}          &                         &                         & \textbf{FSh}          &                       &                       & \textbf{DFDCP}        &                       &                       & \textbf{UADFV}        &                       &                       \\
                & \textbf{ACC}          & \textbf{AUC}          & \textbf{AP}           & \textbf{ACC}          & \textbf{AUC}          & \textbf{AP}           & \textbf{ACC}          & \textbf{AUC}            & \textbf{AP}             & \textbf{ACC}          & \textbf{AUC}          & \textbf{AP}           & \textbf{ACC}          & \textbf{AUC}          & \textbf{AP}           & \textbf{ACC}          & \textbf{AUC}          & \textbf{AP}           \\
Source          & 0.6171           & 0.5730           & 0.6797           & 0.6621           & 0.6118           & 0.7337           & 0.8337           & 0.5570             & 0.8891             & \textbf{0.5370}   & 0.5587           & 0.5480           & 0.6737           & 0.6553           & 0.7598           & 0.6316           & 0.7109           & 0.6443           \\
TENT            & 0.6334           & 0.6166           & 0.7028           & 0.6370           & 0.6327           & 0.7475           & 0.7631           & 0.6409             & 0.9258             & 0.5285           & 0.5586           & 0.5540           & 0.7213           & 0.6990           & 0.7763           & 0.6625           & 0.7330           & 0.6674           \\
MEMO            & 0.6456           & 0.6216           & 0.7003           & 0.6679           & 0.5937           & 0.7171           & 0.8798           & 0.5884             & 0.9148             & 0.5107           & 0.5619           & 0.5408           & 0.7000            & 0.6892           & 0.7466           & 0.6337           & 0.7295           & 0.6653           \\
EATA            & 0.6313           & 0.6165           & 0.7029           & 0.6389           & 0.6330           & 0.7474           & 0.7579           & 0.6438    & 0.9276             & 0.5307           & 0.5583           & 0.5532           & 0.7245           & 0.7004           & 0.7758           & 0.6604           & 0.7330           & 0.6685           \\
CoTTA           & 0.6354           & 0.6280           & 0.6975           & 0.6602           & 0.6189           & 0.7380          & 0.8757           & 0.6068             & 0.9222             & 0.5292           & 0.5661           & 0.5528           & 0.6934           & 0.6524           & 0.7384           & 0.6316           & 0.7210           & 0.6532           \\
LAME            & 0.6211         & 0.5901         & 0.6733         & 0.6505         & 0.5914         & 0.7033         & 0.8935         & 0.5724           & 0.9091           & 0.5007         & 0.5307         & 0.5174         & 0.6475         & 0.5988         & 0.6996         & 0.5102         & 0.676        & 0.6284        \\
VIDA            & 0.6374           & 0.6057           & 0.6683           & \textbf{0.6756}  & 0.5589           & 0.6849           & \textbf{0.8810}  & 0.5948             & 0.9230             & 0.5192           & 0.5285           & 0.5337           & 0.6770           & 0.6925           & 0.7692           & 0.6090           & 0.6972           & 0.6149           \\
COME            & 0.6334         & 0.6162         & 0.7041         & 0.6389         & 0.6327         & 0.7465         & 0.7573         & \textbf{0.6451}  & \textbf{0.9286}  & 0.5292         & 0.5585         & 0.5537         & 0.7262         & 0.7013         & 0.7764         & 0.6625         & 0.7317         & 0.6674         \\
\texttt{T$^2$A} (Ours)      & \textbf{0.6700}  & \textbf{0.6748}  & \textbf{0.7299}  & 0.6718           & \textbf{0.6430}  & \textbf{0.7565}  & 0.7594           & 0.6438             & 0.9279             &  \textbf{0.5370}  & \textbf{0.5728}  & \textbf{0.5657}  & \textbf{0.7327}  & \textbf{0.7320}  & \textbf{0.7774}  & \textbf{0.6830}  & \textbf{0.7623}  & \textbf{0.7117}  
\end{tblr}
\end{table*}

\section{Experiments}\label{sec:experiment}
In this section, we demonstrate the effectiveness of our \texttt{T$^2$A} method when comparing it with state-of-the-art (SoTA) TTA  approaches and DF detectors. 
We also provide an ablation study for our method in Appendix \ref{subsec:ablation-study} and an analysis of running time compared to other TTA methods in Appendix \ref{subsec:running-time}.

\subsection{Setup}

\subsubsection{Datasets and modeling}
We use Xception \cite{chollet2017xception} as the source model, which as commonly used as the backbone in DF detectors. The training set is FaceForensics++ (FF++) \cite{rossler2019faceforensics++}.
To evaluate the adaptability of our \texttt{T$^2$A} method, we use six more datasets at inference time, including CelebDF-v1 \cite{li2020celeb}, CelebDF-v2 \cite{li2020celeb}, DeepFakeDetection (DFD) \cite{DFD}, DeepFake Detection Challenge Preview (DFDCP) \cite{dolhansky2019dee}, UADFV \cite{li2018ictu}, and FaceShifter (FSh) \cite{li2020advancing}. The dataset implementations are provided by \cite{yan2023deepfakebench} and more details are described in Appendix \ref{sec:experiment-details}.

\subsubsection{Metrics}
We use three evaluation metrics: accuracy (ACC), the area under the ROC curve (AUC), and average precision (AP). For each metric, higher values show better results. Notably, in the DF detection context, datasets inherently exhibit significant class imbalance with fake samples substantially dominating real ones \cite{layton2024sok}, the AUC metric is more important as it remains robust to this problem.

\subsubsection{Postprocessing Techniques}
Following \cite{chen2022ost}, we employ four postprocessing techniques: Gaussian blur, changes in color saturation, changes in color contrast, and resize: downsample the image by a factor then upsample it to the original resolution. At the inference time, test samples are applied to these operations with the intensity level increasing from $1$ to $5$. Details of postprocessing techniques and intensity levels are provided in Appendix \ref{sec:experiment-details}. Note that these postprocessing techniques are unknown to all models. 

\definecolor{BlackSqueeze}{rgb}{0.866,0.913,0.949}
\begin{table*}[!ht]
\centering
\fontsize{6pt}{6pt}\selectfont
\caption{Improvement of deepfake detectors to unknown postprocessing techniques. All these methods undergo five levels of intensity of postprocessing techniques.}
\label{tab:DF-post-process}
\begin{tblr}{
  width = \linewidth,
  colspec = {Q[70]Q[30]Q[30]Q[30]Q[30]Q[30]Q[30]Q[30]Q[30]Q[30]Q[30]Q[30]Q[30]Q[30]Q[30]Q[30]},
  cells = {c},
  row{4} = {BlackSqueeze},
  row{6} = {BlackSqueeze},
  row{8} = {BlackSqueeze},
  row{10} = {BlackSqueeze},
  cell{1}{1} = {r=2}{},
  cell{1}{2} = {c=3}{0.1\linewidth},
  cell{1}{5} = {c=3}{0.1\linewidth},
  cell{1}{8} = {c=3}{0.1\linewidth},
  cell{1}{11} = {c=3}{0.1\linewidth},
  cell{1}{14} = {c=3}{0.1\linewidth},
  vline{2,5,8,11,14} = {1-2}{},
  vline{2,5,8,11,14} = {3-10}{},
  hline{1,11} = {-}{0.08em},
  hline{2} = {2-16}{},
  hline{3} = {-}{},
  hline{3} = {2}{-}{},
}
\textbf{Method} & \textbf{Color Contrast} &                   &                   & \textbf{Color Saturation} &                   &                   & \textbf{Resize}                                 &                   &                   & \textbf{Gaussian Blur} &                   &                   & \textbf{Average}  &                   &                   \\
                & \textbf{ACC}            & \textbf{AUC}      & \textbf{AP}       & \textbf{ACC}              & \textbf{AUC}      & \textbf{AP}       & \textbf{ACC}                                    & \textbf{AUC}      & \textbf{AP}       & \textbf{ACC}           & \textbf{AUC}      & \textbf{AP}       & \textbf{ACC}      & \textbf{AUC}      & \textbf{AP}       \\
CORE            & 0.8154 $\pm$ 0.02       & 0.8245 $\pm$ 0.04 & 0.9349 $\pm$ 0.02 & 0.8237 $\pm$ 0.03         & 0.8067 $\pm$ 0.06 & 0.9395 $\pm$ 0.02 & 0.8360 $\pm$ 0.02                               & 0.8628 $\pm$ 0.03 & 0.9598 $\pm$ 0.01 & 0.8334 $\pm$ 0.02      & 0.8265 $\pm$ 0.05 & 0.9409 $\pm$ 0.02 & 0.8271 $\pm$ 0.01 & 0.830 $\pm$ 0.02  & 0.9438 $\pm$ 0.01 \\
CORE + \texttt{T$^2$A}      & 0.8605 $\pm$ 0.01       & 0.8744 $\pm$ 0.02 & 0.9604 $\pm$ 0.01 & 0.8414 $\pm$ 0.02         & 0.8497 $\pm$ 0.04 & 0.9447 $\pm$ 0.01 & 0.8425 $\pm$ 0.01                               & 0.8897 $\pm$ 0.03 & 0.9511 $\pm$ 0.01 & 0.849 $\pm$ 0.01       & 0.8662 $\pm$ 0.02 & 0.9539 $\pm$ 0.01 & 0.8491 $\pm$ 0.01 & 0.8725 $\pm$ 0.02 & 0.9525 $\pm$ 0.01 \\
Effi.B4         & 0.6980 $\pm$ 0.07       & 0.8464 $\pm$ 0.04 & 0.9531 $\pm$ 0.01 & 0.8491 $\pm$ 0.02         & 0.7973 $\pm$ 0.07 & 0.9262 $\pm$ 0.03 & 0.8314 $\pm$ 0.02                               & 0.8458 $\pm$ 0.04 & 0.9526 $\pm$ 0.01 & 0.8380 $\pm$ 0.02      & 0.7929 $\pm$ 0.06 & 0.9286 $\pm$ 0.03 & 0.8041 $\pm$ 0.02 & 0.8206 $\pm$ 0.02 & 0.9401 $\pm$ 0.01 \\
Effi.B4 + \texttt{T$^2$A}   & 0.8531 $\pm$ 0.02       & 0.8638 $\pm$ 0.02 & 0.9542 $\pm$ 0.01 & 0.8271 $\pm$ 0.03         & 0.8311 $\pm$ 0.05 & 0.9372 $\pm$ 0.02 & 0.8302 $\pm$ 0.02                               & 0.8355 $\pm$ 0.04 & 0.9485 $\pm$ 0.01 & 0.8442 $\pm$ 0.01      & 0.8670 $\pm$ 0.03 & 0.9515 $\pm$ 0.01 & 0.8382 $\pm$ 0.01 & 0.8592 $\pm$ 0.02 & 0.9478 $\pm$ 0.01 \\
F3Net           & 0.8037 $\pm$ 0.03       & 0.8306 $\pm$ 0.05 & 0.9438 $\pm$ 0.02 & 0.8542 $\pm$ 0.02         & 0.8196 $\pm$ 0.07 & 0.9413 $\pm$ 0.02 & 0.8551 $\pm$ 0.03                               & 0.8681 $\pm$ 0.03 & 0.9575 $\pm$ 0.01 & 0.8360 $\pm$ 0.02      & 0.8136 $\pm$ 0.05 & 0.9374 $\pm$ 0.02 & 0.8284 $\pm$ 0.01 & 0.8387 $\pm$ 0.02 & 0.9491 $\pm$ 0.01 \\
F3Net + \texttt{T$^2$A}     & 0.8605 $\pm$ 0.01       & 0.8879 $\pm$ 0.02 & 0.9641 $\pm$ 0.01 & 0.8617 $\pm$ 0.02         & 0.8737 $\pm$ 0.04 & 0.9599 $\pm$ 0.02 & 0.8142 $\pm$ 0.01                               & 0.8723 $\pm$ 0.03 & 0.9632 $\pm$ 0.01 & 0.8417 $\pm$ 0.02      & 0.8489 $\pm$ 0.02 & 0.9524 $\pm$ 0.01 & 0.8547 $\pm$ 0.01 & 0.8776 $\pm$ 0.01 & 0.9621 $\pm$ 0.01 \\
RECCE           & 0.8080 $\pm$ 0.03       & 0.8189 $\pm$ 0.04 & 0.9386 $\pm$ 0.02 & 0.8348 $\pm$ 0.02         & 0.7915 $\pm$ 0.06 & 0.9283 $\pm$ 0.02 & 0.8137 $\pm$ 0.03\textcolor[rgb]{0.8,0.8,0.8}{} & 0.8338 $\pm$ 0.04 & 0.9484 $\pm$ 0.01 & 0.8360 $\pm$ 0.02      & 0.8136 $\pm$ 0.04 & 0.9374 $\pm$ 0.01 & 0.8231 $\pm$ 0.01 & 0.8144 $\pm$ 0.02 & 0.9382 $\pm$ 0.01 \\
RECCE + \texttt{T$^2$A}     & 0.8502 $\pm$ 0.01       & 0.8698 $\pm$ 0.02 & 0.9587 $\pm$ 0.01 & 0.8291 $\pm$ 0.02         & 0.8432 $\pm$ 0.05 & 0.9406 $\pm$ 0.02 & 0.8408 $\pm$ 0.01                               & 0.8426 $\pm$ 0.03 & 0.9495 $\pm$ 0.01 & 0.8417 $\pm$ 0.01      & 0.8689 $\pm$ 0.02 & 0.9524 $\pm$ 0.01 & 0.8405 $\pm$ 0.01 & 0.8561 $\pm$ 0.01 & 0.9503 $\pm$ 0.01 
\end{tblr}
\end{table*}

\definecolor{BlackSqueeze}{rgb}{0.866,0.913,0.949}
\begin{table*}[!ht]
\centering
\fontsize{6pt}{6pt}\selectfont
\caption{Improvement of deepfake detectors to unknown data distributions and postprocessing techniques across six Deepfake datasets.}
\label{tab:DF-cross-dataset}
\begin{tblr}{
  width = \linewidth,
  colspec = {Q[90]Q[30]Q[30]Q[30]Q[30]Q[30]Q[30]Q[30]Q[30]Q[30]Q[30]Q[30]Q[30]Q[30]Q[30]Q[30]Q[30]Q[30]Q[30]},
  cells = {c},
  row{4} = {BlackSqueeze},
  row{6} = {BlackSqueeze},
  row{8} = {BlackSqueeze},
  row{10} = {BlackSqueeze},
  cell{1}{1} = {r=2}{},
  cell{1}{2} = {c=3}{0.1\linewidth},
  cell{1}{5} = {c=3}{0.1\linewidth},
  cell{1}{8} = {c=3}{0.1\linewidth},
  cell{1}{11} = {c=3}{0.1\linewidth},
  cell{1}{14} = {c=3}{0.1\linewidth},
  cell{1}{17} = {c=3}{0.1\linewidth},
  vline{2,5,8,11,14,17} = {1-10}{},
  hline{1,11} = {-}{0.08em},
  hline{2} = {2-19}{},
  hline{3} = {-}{},
  hline{3} = {2}{-}{},
}
\textbf{Method} & \textbf{CelebDF-v1} &              &             & \textbf{CelebDF-v2} &              &             & \textbf{DFD} &              &             & \textbf{FSh} &              &             & \textbf{DFDCP} &              &             & \textbf{UADFV} &              &             \\
                & \textbf{ACC}        & \textbf{AUC} & \textbf{AP} & \textbf{ACC}        & \textbf{AUC} & \textbf{AP} & \textbf{ACC} & \textbf{AUC} & \textbf{AP} & \textbf{ACC} & \textbf{AUC} & \textbf{AP} & \textbf{ACC}   & \textbf{AUC} & \textbf{AP} & \textbf{ACC}   & \textbf{AUC} & \textbf{AP} \\
CORE            & 0.6517              & 0.6828       & 0.7837      &  0.6467             & 0.6268       & 0.7527      & 0.8515       & 0.5319       & 0.8962      & 0.5050        & 0.5216       & 0.5151      & 0.7016         & 0.6465       & 0.7513      & 0.6090         & 0.7481       & 0.7331      \\
CORE + \texttt{T$^2$A}      & 0.6558              & 0.6883       & 0.7599      & 0.7162              & 0.6571       & 0.7576      & 0.7946       & 0.6292       & 0.9291      & 0.5200       & 0.5103       & 0.4985      & 0.6721         & 0.6611       & 0.7565      & 0.6337         & 0.7805       & 0.7692      \\
Effi.B4         & 0.6313              & 0.6613       & 0.7202      & 0.6428              & 0.5489       & 0.6556      & 0.8743       & 0.6310        & 0.9282      & 0.5292       & 0.5737       & 0.5504      & 0.6344         & 0.5023       & 0.6438      & 0.5576         & 0.6791       & 0.6363      \\
Effi.B4 + \texttt{T$^2$A}   & 0.6415              & 0.6659       & 0.7542      & 0.6351              & 0.4347       & 0.7312      & 0.8259       & 0.6892       & 0.9452      & 0.5450       & 0.5944       & 0.5598      & 0.6475         & 0.5824       & 0.7040      & 0.6152         & 0.7107       & 0.6622      \\
F3Net           & 0.6252              & 0.6541       & 0.7614      & 0.6563              & 0.6604       & 0.7681      & 0.8547       & 0.5507       & 0.9012      & 0.5228       & 0.5448       & 0.5644      & 0.6688         & 0.6528       & 0.7443      & 0.5843         & 0.7146       & 0.6866      \\
F3Net + \texttt{T$^2$A}     & 0.6517              & 0.6655       & 0.7531      &  0.6602             & 0.6409       & 0.7283      & 0.7500       & 0.6097       & 0.9244      & 0.5128       & 0.5569       & 0.5647      & 0.6803         & 0.6961       & 0.7831      & 0.6563         & 0.7447       & 0.6877      \\
RECCE           & 0.5804              & 0.5689       & 0.6804      &  0.6776             & 0.6175       & 0.7531      & 0.8177       & 0.6256       & 0.9356      & 0.5235       & 0.5367       & 0.5275      & 0.6672         & 0.6358       & 0.7333      & 0.6522         & 0.7194       & 0.6778      \\
RECCE + \texttt{T$^2$A}     & 0.6578              & 0.6508       & 0.7233      & 0.6718              & 0.6725       & 0.7783      & 0.7296       & 0.6521       & 0.9346      & 0.5321       & 0.5512       & 0.5593      & 0.7032         & 0.7184       & 0.7949      & 0.7119         & 0.7910        & 0.7370       
\end{tblr}
\end{table*}

\subsubsection{Baselines}
For TTA, we compare our \texttt{T$^2$A} method with SOTA methods, including TENT \cite{wang2020tent}, MEMO \cite{zhang2022memo}, EATA \cite{niu2022efficient}, CoTTA \cite{wang2022continual}, LAME \cite{boudiaf2022parameter}, ViDA \cite{liuvida2023}, and COME \cite{zhang2024come}. For DF detection, we employ the following DF detectors: EfficientNetB4 \cite{tan2019efficientnet}, F3Net \cite{qian2020thinking}, CORE \cite{ni2022core}, RECCE \cite{cao2022end}. Details for these baselines are provided in Appendix \ref{sec:experiment-details}.

\subsubsection{Implementation}
For adaptation, we use Adam optimizer with learning rate $\eta = 1e-4$, batch size of 32. Other hyperparameters including loss balancing ones $\alpha, \beta$ and gradient masking threshold $\psi$ are selected by a grid-search manner from defined values in Table \ref{tab:hyperparams}. 
The $\gamma$ hyperparameter in Eq. \ref{eq:focal_loss} is set to $2.0$. Details about these hyperparameters are provided in SAppendix \ref{ssec:postprocessing-intensity}. 

\subsection{Experimental Results}
We design the experiments to assess the effectiveness of our method under two real-world scenarios: (i) unknown postprocessing techniques, and (ii) both unknown data distributions and postprocessing techniques. The primary distinction between these scenarios lies in the underlying data distribution assumptions. In the first scenario, we assume that test samples are drawn from a distribution similar to the training data and focus specifically on evaluating our method's resilience when adversaries intentionally employ unknown postprocessing techniques. The second scenario presents a more challenging setting where test samples stem from unknown distributions, allowing us to evaluate not only the method's resilience to postprocessing techniques but also its broader generalization across different data domains. 

\subsubsection{Comparison with SoTA TTA Approaches}
We compare our \texttt{T$^2$A} method with existing TTA approaches, with results presented in Table \ref{tab:TTA-post-process} and Table \ref{tab:TTA-cross-dataset}. 
%
Table \ref{tab:TTA-post-process} reports results when tested with unknown postprocessing techniques. 
Each technique is tested across five intensity levels, with the results showing averaged performance metrics. Detailed results for individual intensity levels are provided in Appendix \ref{sec:more-results}.
The \textit{Average} column denotes the mean across all postprocessing techniques, providing a holistic view of adaptation capability. 
We test our method and other TTA approaches on FF++ samples exposed to unseen postprocessing operations.
From Table \ref{tab:TTA-post-process}, we can observe that our method outperforms existing TTA approaches. On average, \texttt{T$^2$A} improves the source DF detector by $2.93 \%$ on AUC.
For the more challenging scenario - unknown data distributions and postprocessing techniques, Table \ref{tab:TTA-cross-dataset} shows that \texttt{T$^2$A} achieves SoTA results on $5$ out of $6$ datasets, including CelebDF-1, CelebDF-2, FSh, DFDCP, and UADFV, and the second-best result on DFD dataset. Note that postprocessing techniques used in this experiment are unseen during the training process of the source model. 

\subsubsection{Adaptability Improvement over Deepfake Detectors}
To further demonstrate the effectiveness of our \texttt{T$^2$A} method, we evaluate its capability to enhance the adaptability of DF detectors.
We test the performance of DF detectors with and without the \texttt{T$^2$A} method under two scenarios. 
%
For the first scenario, Table \ref{tab:DF-post-process} indicates that:  When integrated with \texttt{T$^2$A}, the performance of DF detectors measured by AUC is significantly improved, enhancing the resilience of these detectors against unseen postprocessing techniques. Particularly, our method shows substantial improvements of $4.25\%$ for CORE, $3.86\%$ for EfficientNet-B4, $3.89\%$ for F3Net, and $4.17\%$ for RECCE. 
Under the more challenging scenario, Table \ref{tab:DF-cross-dataset} presents results that \texttt{T$^2$A} consistently enhances the generalization capability of DF detectors over unseen data distributions while maintaining robustness against postprocessing manipulations. For example, on the real-world DF benchmark DFDCP, our method improves the performance of RECCE to $8.26\%$, EfficientNet-B4 to $8\%$, F3Net to $4.33\%$, and CORE to $1.46\%$. 





\section{Conclusion}
In this work, we introduce \texttt{T$^2$A}, which improves the adaptability of DF detectors across two challenging scenarios: unknown postprocessing techniques and data distributions during inference time. Instead of solely relying on EM, \texttt{T$^2$A} enables the model to explore alternative options before decision-making through NL with noisy pseudo-labels. We also provide a theoretical analysis to demonstrate that the proposed objective exhibits complementary behavior to EM. Through experiments, we show that \texttt{T$^2$A} achieves higher adaptation performance compared to SoTA TTA approaches. Furthermore, when integrated with \texttt{T$^2$A}, the resilience and generalization of DF detectors can be significantly improved without requiring additional training data or architectural modifications, making it particularly valuable for real-world deployments. However, since our method is based on backpropagation for updating parameters at inference time, it only works with end-to-end DF detectors that allow gradient flow throughout the model.

\section*{Acknowledgments}
This publication has emanated from research conducted with the financial support of Science Foundation Ireland under Grant number 18/CRT/6183.


\bibliographystyle{named}
\bibliography{ijcai25}

\begin{thebibliography}{}

\bibitem[\protect\citeauthoryear{Boudiaf \bgroup \em et al.\egroup }{2022}]{boudiaf2022parameter}
Malik Boudiaf, Romain Mueller, Ismail Ben~Ayed, and Luca Bertinetto.
\newblock Parameter-free online test-time adaptation.
\newblock In {\em Proceedings of the IEEE/CVF Conference on Computer Vision and Pattern Recognition}, pages 8344--8353, 2022.

\bibitem[\protect\citeauthoryear{Cao \bgroup \em et al.\egroup }{2022}]{cao2022end}
Junyi Cao, Chao Ma, Taiping Yao, Shen Chen, Shouhong Ding, and Xiaokang Yang.
\newblock End-to-end reconstruction-classification learning for face forgery detection.
\newblock In {\em Proceedings of the IEEE/CVF Conference on Computer Vision and Pattern Recognition}, pages 4113--4122, 2022.

\bibitem[\protect\citeauthoryear{Chen \bgroup \em et al.\egroup }{2022}]{chen2022ost}
Liang Chen, Yong Zhang, Yibing Song, Jue Wang, and Lingqiao Liu.
\newblock Ost: Improving generalization of deepfake detection via one-shot test-time training.
\newblock {\em Advances in Neural Information Processing Systems}, 35:24597--24610, 2022.

\bibitem[\protect\citeauthoryear{Chollet}{2017}]{chollet2017xception}
Fran{\c{c}}ois Chollet.
\newblock Xception: Deep learning with depthwise separable convolutions.
\newblock In {\em Proceedings of the IEEE conference on computer vision and pattern recognition}, pages 1251--1258, 2017.

\bibitem[\protect\citeauthoryear{Corvi \bgroup \em et al.\egroup }{2023}]{corvi2023intriguing}
Riccardo Corvi, Davide Cozzolino, Giovanni Poggi, Koki Nagano, and Luisa Verdoliva.
\newblock Intriguing properties of synthetic images: from generative adversarial networks to diffusion models.
\newblock In {\em Proceedings of the IEEE/CVF Conference on Computer Vision and Pattern Recognition}, pages 973--982, 2023.

\bibitem[\protect\citeauthoryear{Dolhansky}{2019}]{dolhansky2019dee}
B~Dolhansky.
\newblock The deepfake detection challenge (dfdc) preview dataset.
\newblock {\em arXiv preprint arXiv:1910.08854}, 2019.

\bibitem[\protect\citeauthoryear{Fang \bgroup \em et al.\egroup }{2024}]{fang2024unified}
Hao Fang, Ajian Liu, Haocheng Yuan, Junze Zheng, Dingheng Zeng, Yanhong Liu, Jiankang Deng, Sergio Escalera, Xiaoming Liu, Jun Wan, et~al.
\newblock Unified physical-digital face attack detection.
\newblock In {\em Proceedings of the Thirty-Third International Joint Conference on Artificial Intelligence}, pages 749--757, 2024.

\bibitem[\protect\citeauthoryear{Frank \bgroup \em et al.\egroup }{2020}]{frank2020leveraging}
Joel Frank, Thorsten Eisenhofer, Lea Sch{\"o}nherr, Asja Fischer, Dorothea Kolossa, and Thorsten Holz.
\newblock Leveraging frequency analysis for deepfake image recognition.
\newblock In {\em International conference on machine learning}, pages 3247--3258. PMLR, 2020.

\bibitem[\protect\citeauthoryear{Ghosh \bgroup \em et al.\egroup }{2017}]{ghosh2017robust}
Aritra Ghosh, Himanshu Kumar, and P~Shanti Sastry.
\newblock Robust loss functions under label noise for deep neural networks.
\newblock In {\em Proceedings of the AAAI conference on artificial intelligence}, volume~31, 2017.

\bibitem[\protect\citeauthoryear{Google}{2019}]{DFD}
Google.
\newblock Contributing data to deepfake detection research, 2019.
\newblock Accessed on 11 December 2024.

\bibitem[\protect\citeauthoryear{He \bgroup \em et al.\egroup }{2024}]{he2024joint}
Xianhua He, Dashuang Liang, Song Yang, Zhanlong Hao, Hui Ma, Binjie Mao, Xi~Li, Yao Wang, Pengfei Yan, and Ajian Liu.
\newblock Joint physical-digital facial attack detection via simulating spoofing clues.
\newblock In {\em Proceedings of the IEEE/CVF Conference on Computer Vision and Pattern Recognition}, pages 995--1004, 2024.

\bibitem[\protect\citeauthoryear{Hendrycks and Dietterich}{2019}]{hendrycks2019benchmarking}
Dan Hendrycks and Thomas Dietterich.
\newblock Benchmarking neural network robustness to common corruptions and perturbations.
\newblock {\em International Conference on Learning Representations}, 2019.

\bibitem[\protect\citeauthoryear{Kim \bgroup \em et al.\egroup }{2019}]{kim2019nlnl}
Youngdong Kim, Junho Yim, Juseung Yun, and Junmo Kim.
\newblock Nlnl: Negative learning for noisy labels.
\newblock In {\em Proceedings of the IEEE/CVF international conference on computer vision}, pages 101--110, 2019.

\bibitem[\protect\citeauthoryear{Kim \bgroup \em et al.\egroup }{2021}]{kim2021joint}
Youngdong Kim, Juseung Yun, Hyounguk Shon, and Junmo Kim.
\newblock Joint negative and positive learning for noisy labels.
\newblock In {\em Proceedings of the IEEE/CVF conference on computer vision and pattern recognition}, pages 9442--9451, 2021.

\bibitem[\protect\citeauthoryear{Layton \bgroup \em et al.\egroup }{2024}]{layton2024sok}
Seth Layton, Tyler Tucker, Daniel Olszewski, Kevin Warren, Kevin Butler, and Patrick Traynor.
\newblock $\{$SoK$\}$: The good, the bad, and the unbalanced: Measuring structural limitations of deepfake media datasets.
\newblock In {\em 33rd USENIX Security Symposium (USENIX Security 24)}, pages 1027--1044, 2024.

\bibitem[\protect\citeauthoryear{Li \bgroup \em et al.\egroup }{2018}]{li2018ictu}
Yuezun Li, Ming-Ching Chang, and Siwei Lyu.
\newblock In ictu oculi: Exposing ai created fake videos by detecting eye blinking.
\newblock In {\em 2018 IEEE International workshop on information forensics and security (WIFS)}, pages 1--7. Ieee, 2018.

\bibitem[\protect\citeauthoryear{Li \bgroup \em et al.\egroup }{2020a}]{li2020advancing}
Lingzhi Li, Jianmin Bao, Hao Yang, Dong Chen, and Fang Wen.
\newblock Advancing high fidelity identity swapping for forgery detection.
\newblock In {\em Proceedings of the IEEE/CVF conference on computer vision and pattern recognition}, pages 5074--5083, 2020.

\bibitem[\protect\citeauthoryear{Li \bgroup \em et al.\egroup }{2020b}]{li2020celeb}
Yuezun Li, Xin Yang, Pu~Sun, Honggang Qi, and Siwei Lyu.
\newblock Celeb-df: A large-scale challenging dataset for deepfake forensics.
\newblock In {\em Proceedings of the IEEE/CVF conference on computer vision and pattern recognition}, pages 3207--3216, 2020.

\bibitem[\protect\citeauthoryear{Li \bgroup \em et al.\egroup }{2024}]{li2024comprehensive}
Jingjing Li, Zhiqi Yu, Zhekai Du, Lei Zhu, and Heng~Tao Shen.
\newblock A comprehensive survey on source-free domain adaptation.
\newblock {\em IEEE Transactions on Pattern Analysis and Machine Intelligence}, 2024.

\bibitem[\protect\citeauthoryear{Liang \bgroup \em et al.\egroup }{2024}]{liang2024comprehensive}
Jian Liang, Ran He, and Tieniu Tan.
\newblock A comprehensive survey on test-time adaptation under distribution shifts.
\newblock {\em International Journal of Computer Vision}, pages 1--34, 2024.

\bibitem[\protect\citeauthoryear{Liu \bgroup \em et al.\egroup }{2021}]{liu2021spatial}
Honggu Liu, Xiaodan Li, Wenbo Zhou, Yuefeng Chen, Yuan He, Hui Xue, Weiming Zhang, and Nenghai Yu.
\newblock Spatial-phase shallow learning: rethinking face forgery detection in frequency domain.
\newblock In {\em Proceedings of the IEEE/CVF conference on computer vision and pattern recognition}, pages 772--781, 2021.

\bibitem[\protect\citeauthoryear{Liu \bgroup \em et al.\egroup }{2023a}]{liuvida2023}
Jiaming Liu, Senqiao Yang, Peidong Jia, Renrui Zhang, Ming Lu, Yandong Guo, Wei Xue, and Shanghang Zhang.
\newblock Vida: Homeostatic visual domain adapter for continual test time adaptation.
\newblock In {\em International Conference on Learning Representations}, 2023.

\bibitem[\protect\citeauthoryear{Liu \bgroup \em et al.\egroup }{2023b}]{liu2023adaptive}
Jiawei Liu, Jingyi Xie, Yang Wang, and Zheng-Jun Zha.
\newblock Adaptive texture and spectrum clue mining for generalizable face forgery detection.
\newblock {\em IEEE Transactions on Information Forensics and Security}, 2023.

\bibitem[\protect\citeauthoryear{Liu \bgroup \em et al.\egroup }{2024}]{liu2024cfpl}
Ajian Liu, Shuai Xue, Jianwen Gan, Jun Wan, Yanyan Liang, Jiankang Deng, Sergio Escalera, and Zhen Lei.
\newblock Cfpl-fas: Class free prompt learning for generalizable face anti-spoofing.
\newblock In {\em Proceedings of the IEEE/CVF Conference on Computer Vision and Pattern Recognition}, pages 222--232, 2024.

\bibitem[\protect\citeauthoryear{Ma \bgroup \em et al.\egroup }{2020}]{ma2020normalized}
Xingjun Ma, Hanxun Huang, Yisen Wang, Simone Romano, Sarah Erfani, and James Bailey.
\newblock Normalized loss functions for deep learning with noisy labels.
\newblock In {\em International conference on machine learning}, pages 6543--6553. PMLR, 2020.

\bibitem[\protect\citeauthoryear{Nguyen-Le \bgroup \em et al.\egroup }{2024a}]{nguyen2024deepfake}
Hong-Hanh Nguyen-Le, Van-Tuan Tran, Dinh-Thuc Nguyen, and Nhien-An Le-Khac.
\newblock Deepfake generation and proactive deepfake defense: A comprehensive survey.
\newblock {\em Authorea Preprints}, 2024.

\bibitem[\protect\citeauthoryear{Nguyen-Le \bgroup \em et al.\egroup }{2024b}]{nguyen2024passive}
Hong-Hanh Nguyen-Le, Van-Tuan Tran, Dinh-Thuc Nguyen, and Nhien-An Le-Khac.
\newblock Passive deepfake detection across multi-modalities: A comprehensive survey.
\newblock {\em arXiv preprint arXiv:2411.17911}, 2024.

\bibitem[\protect\citeauthoryear{Ni \bgroup \em et al.\egroup }{2022}]{ni2022core}
Yunsheng Ni, Depu Meng, Changqian Yu, Chengbin Quan, Dongchun Ren, and Youjian Zhao.
\newblock Core: Consistent representation learning for face forgery detection.
\newblock In {\em Proceedings of the IEEE/CVF conference on computer vision and pattern recognition}, pages 12--21, 2022.

\bibitem[\protect\citeauthoryear{Niu \bgroup \em et al.\egroup }{2022}]{niu2022efficient}
Shuaicheng Niu, Jiaxiang Wu, Yifan Zhang, Yaofo Chen, Shijian Zheng, Peilin Zhao, and Mingkui Tan.
\newblock Efficient test-time model adaptation without forgetting.
\newblock In {\em International conference on machine learning}, pages 16888--16905. PMLR, 2022.

\bibitem[\protect\citeauthoryear{Niu \bgroup \em et al.\egroup }{2023}]{niu2023towards}
Shuaicheng Niu, Jiaxiang Wu, Yifan Zhang, Zhiquan Wen, Yaofo Chen, Peilin Zhao, and Mingkui Tan.
\newblock Towards stable test-time adaptation in dynamic wild world.
\newblock {\em The Eleventh International Conference on Learning Representations}, 2023.

\bibitem[\protect\citeauthoryear{Ojha \bgroup \em et al.\egroup }{2023}]{ojha2023towards}
Utkarsh Ojha, Yuheng Li, and Yong~Jae Lee.
\newblock Towards universal fake image detectors that generalize across generative models.
\newblock In {\em Proceedings of the IEEE/CVF Conference on Computer Vision and Pattern Recognition}, pages 24480--24489, 2023.

\bibitem[\protect\citeauthoryear{Pan \bgroup \em et al.\egroup }{2023}]{pan2023dfil}
Kun Pan, Yifang Yin, Yao Wei, Feng Lin, Zhongjie Ba, Zhenguang Liu, Zhibo Wang, Lorenzo Cavallaro, and Kui Ren.
\newblock Dfil: Deepfake incremental learning by exploiting domain-invariant forgery clues.
\newblock In {\em Proceedings of the 31st ACM International Conference on Multimedia}, pages 8035--8046, 2023.

\bibitem[\protect\citeauthoryear{Qian \bgroup \em et al.\egroup }{2020}]{qian2020thinking}
Yuyang Qian, Guojun Yin, Lu~Sheng, Zixuan Chen, and Jing Shao.
\newblock Thinking in frequency: Face forgery detection by mining frequency-aware clues.
\newblock In {\em European conference on computer vision}, pages 86--103. Springer, 2020.

\bibitem[\protect\citeauthoryear{Ross and Doll{\'a}r}{2017}]{ross2017focal}
T-YLPG Ross and GKHP Doll{\'a}r.
\newblock Focal loss for dense object detection.
\newblock In {\em proceedings of the IEEE conference on computer vision and pattern recognition}, pages 2980--2988, 2017.

\bibitem[\protect\citeauthoryear{Rossler \bgroup \em et al.\egroup }{2019}]{rossler2019faceforensics++}
Andreas Rossler, Davide Cozzolino, Luisa Verdoliva, Christian Riess, Justus Thies, and Matthias Nie{\ss}ner.
\newblock Faceforensics++: Learning to detect manipulated facial images.
\newblock In {\em Proceedings of the IEEE/CVF international conference on computer vision}, pages 1--11, 2019.

\bibitem[\protect\citeauthoryear{Schneider \bgroup \em et al.\egroup }{2020}]{schneider2020improving}
Steffen Schneider, Evgenia Rusak, Luisa Eck, Oliver Bringmann, Wieland Brendel, and Matthias Bethge.
\newblock Improving robustness against common corruptions by covariate shift adaptation.
\newblock {\em Advances in neural information processing systems}, 33:11539--11551, 2020.

\bibitem[\protect\citeauthoryear{Shiohara and Yamasaki}{2022}]{shiohara2022detecting}
Kaede Shiohara and Toshihiko Yamasaki.
\newblock Detecting deepfakes with self-blended images.
\newblock In {\em Proceedings of the IEEE/CVF Conference on Computer Vision and Pattern Recognition}, pages 18720--18729, 2022.

\bibitem[\protect\citeauthoryear{Tan and Le}{2019}]{tan2019efficientnet}
Mingxing Tan and Quoc Le.
\newblock Efficientnet: Rethinking model scaling for convolutional neural networks.
\newblock In {\em International conference on machine learning}, pages 6105--6114. PMLR, 2019.

\bibitem[\protect\citeauthoryear{Wang \bgroup \em et al.\egroup }{2020}]{wang2020tent}
Dequan Wang, Evan Shelhamer, Shaoteng Liu, Bruno Olshausen, and Trevor Darrell.
\newblock Tent: Fully test-time adaptation by entropy minimization.
\newblock {\em arXiv preprint arXiv:2006.10726}, 2020.

\bibitem[\protect\citeauthoryear{Wang \bgroup \em et al.\egroup }{2022}]{wang2022continual}
Qin Wang, Olga Fink, Luc Van~Gool, and Dengxin Dai.
\newblock Continual test-time domain adaptation.
\newblock In {\em Proceedings of the IEEE/CVF Conference on Computer Vision and Pattern Recognition}, pages 7201--7211, 2022.

\bibitem[\protect\citeauthoryear{Yan \bgroup \em et al.\egroup }{2023}]{yan2023deepfakebench}
Zhiyuan Yan, Yong Zhang, Xinhang Yuan, Siwei Lyu, and Baoyuan Wu.
\newblock Deepfakebench: A comprehensive benchmark of deepfake detection.
\newblock {\em arXiv preprint arXiv:2307.01426}, 2023.

\bibitem[\protect\citeauthoryear{Yan \bgroup \em et al.\egroup }{2024}]{yan2024transcending}
Zhiyuan Yan, Yuhao Luo, Siwei Lyu, Qingshan Liu, and Baoyuan Wu.
\newblock Transcending forgery specificity with latent space augmentation for generalizable deepfake detection.
\newblock In {\em Proceedings of the IEEE/CVF Conference on Computer Vision and Pattern Recognition}, pages 8984--8994, 2024.

\bibitem[\protect\citeauthoryear{Ye \bgroup \em et al.\egroup }{2023}]{ye2023active}
Xichen Ye, Xiaoqiang Li, Tong Liu, Yan Sun, Weiqin Tong, et~al.
\newblock Active negative loss functions for learning with noisy labels.
\newblock {\em Advances in Neural Information Processing Systems}, 36:6917--6940, 2023.

\bibitem[\protect\citeauthoryear{Zhang \bgroup \em et al.\egroup }{2022}]{zhang2022memo}
Marvin Zhang, Sergey Levine, and Chelsea Finn.
\newblock Memo: Test time robustness via adaptation and augmentation.
\newblock {\em Advances in neural information processing systems}, 35:38629--38642, 2022.

\bibitem[\protect\citeauthoryear{Zhang \bgroup \em et al.\egroup }{2024}]{zhang2024come}
Qingyang Zhang, Yatao Bian, Xinke Kong, Peilin Zhao, and Changqing Zhang.
\newblock Come: Test-time adaption by conservatively minimizing entropy.
\newblock {\em arXiv preprint arXiv:2410.10894}, 2024.

\bibitem[\protect\citeauthoryear{Zhou \bgroup \em et al.\egroup }{2021}]{zhou2021asymmetric}
Xiong Zhou, Xianming Liu, Junjun Jiang, Xin Gao, and Xiangyang Ji.
\newblock Asymmetric loss functions for learning with noisy labels.
\newblock In {\em International conference on machine learning}, pages 12846--12856. PMLR, 2021.

\end{thebibliography}

\clearpage
\appendix

\section*{Appendix for "Think Twice before Adaptation: Improving Adaptability of DeepFake Detection via Online Test-Time Adaptation"}\label{sec:appendix}

\begin{figure*}[ht]
    \centering
    \includegraphics[width=0.98\textwidth]{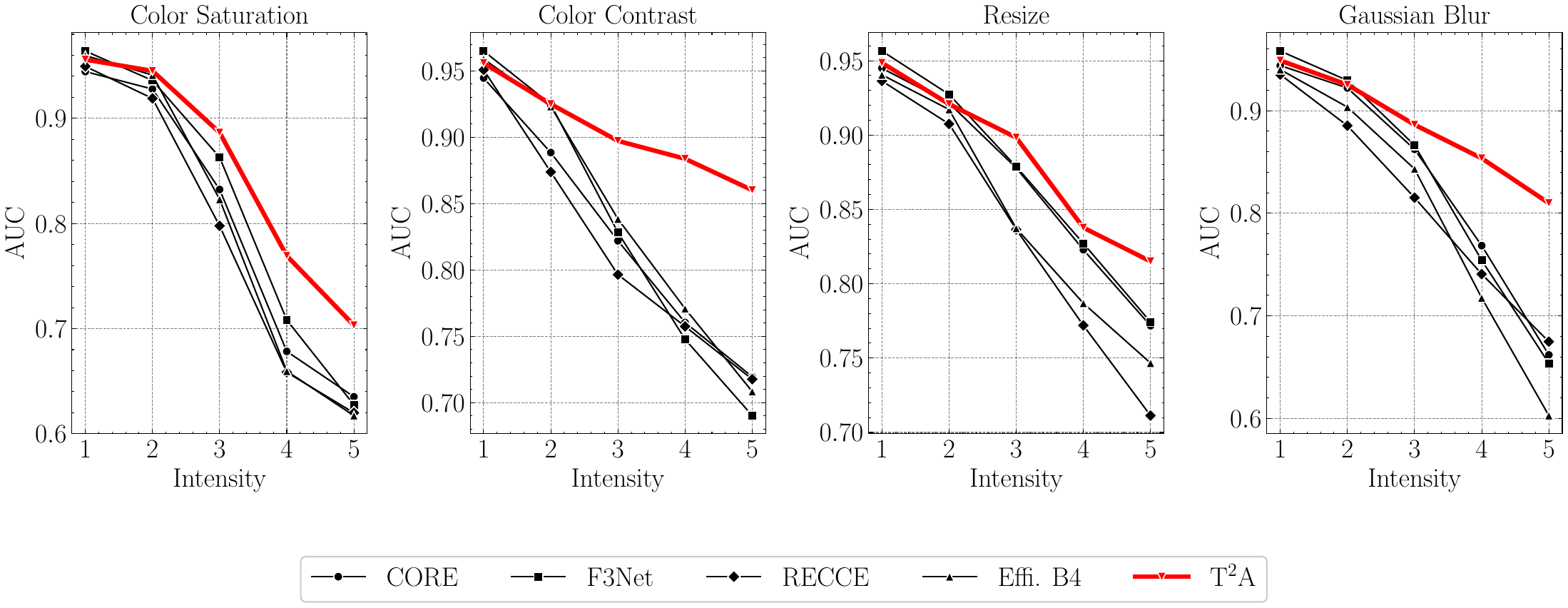}
    \caption{Resilience capability comparisons of different DF detectors and our method under various unknown postprocessing techniques, including color saturation, color contrast, downsampling, and Gaussian blurring. The results are aggregated across five intensity levels. All these methods undergo 5 levels of intensity of postprocessing techniques.}
    \label{fig:exp_cross_augs}
\end{figure*}

\begin{figure*}[!ht]
    \centering
    \fontsize{7pt}{7pt}\selectfont
    \begin{tblr}{
        colspec = {Q[m,3cm]Q[c,45]Q[c,45]Q[c,45]Q[c,45]Q[c,45]},
        column{1} = {halign=r},
    }
        & Intensity 1 & Intensity 2 & Intensity 3 & Intensity 4 & Intensity 5 \\
        
        {Gaussian Blur} & 
        \includegraphics[width=0.15\textwidth]{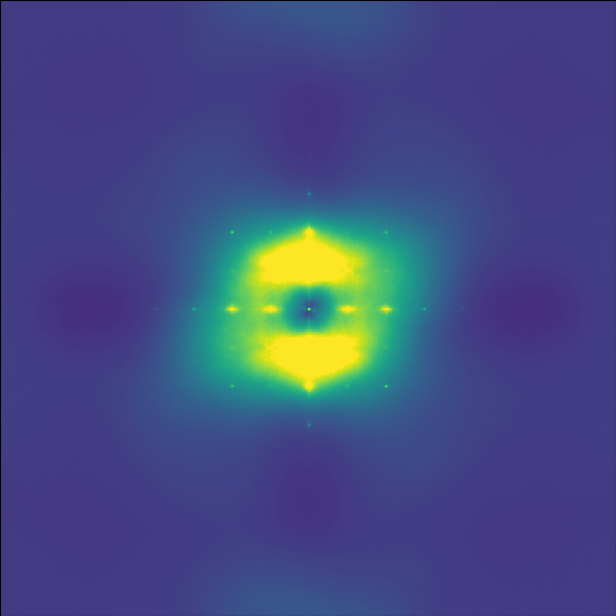} &
        \includegraphics[width=0.15\textwidth]{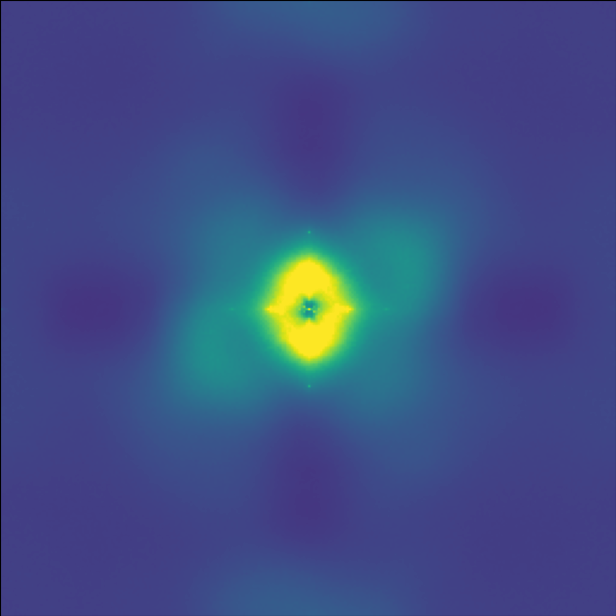} &
        \includegraphics[width=0.15\textwidth]{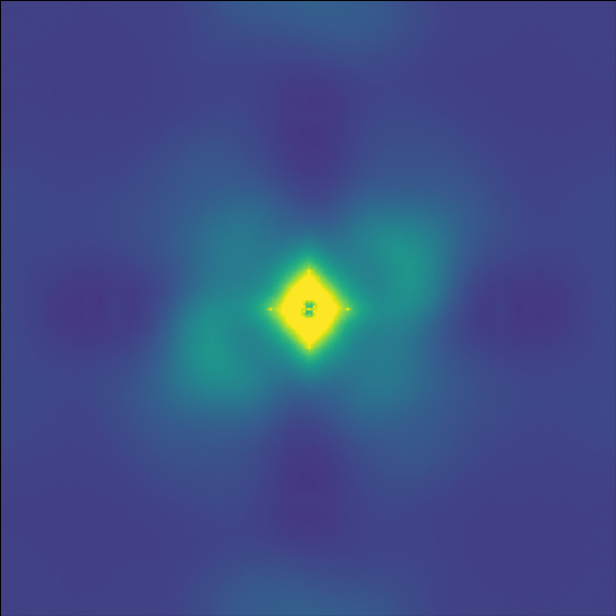} &
        \includegraphics[width=0.15\textwidth]{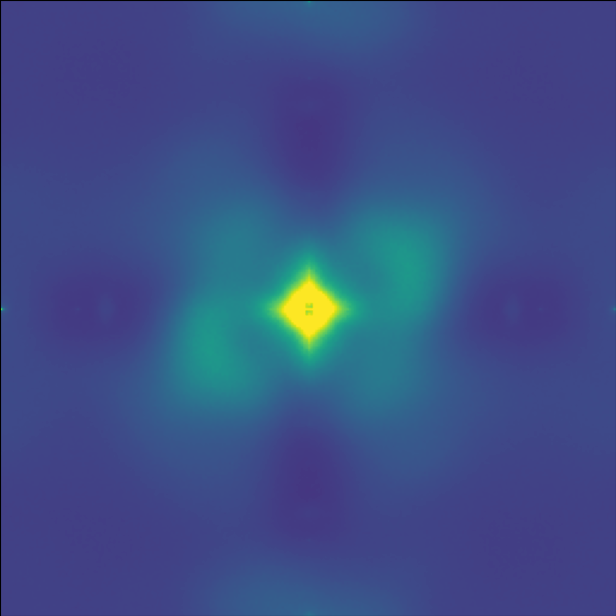} &
        \includegraphics[width=0.15\textwidth]{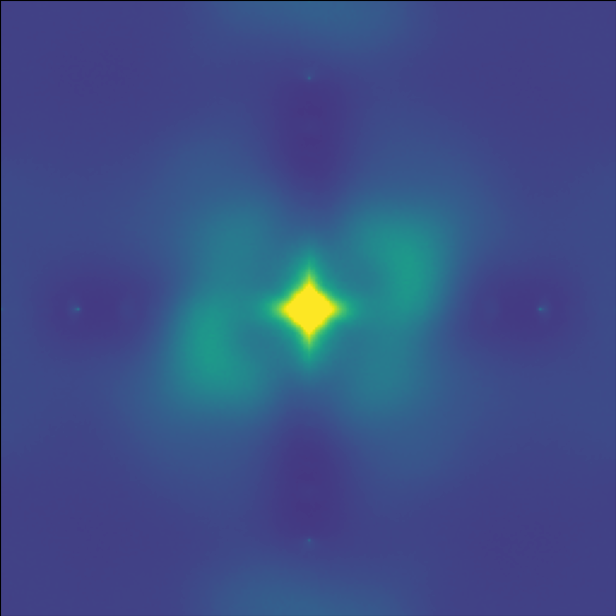} \\
        
        Resize &
        \includegraphics[width=0.15\textwidth]{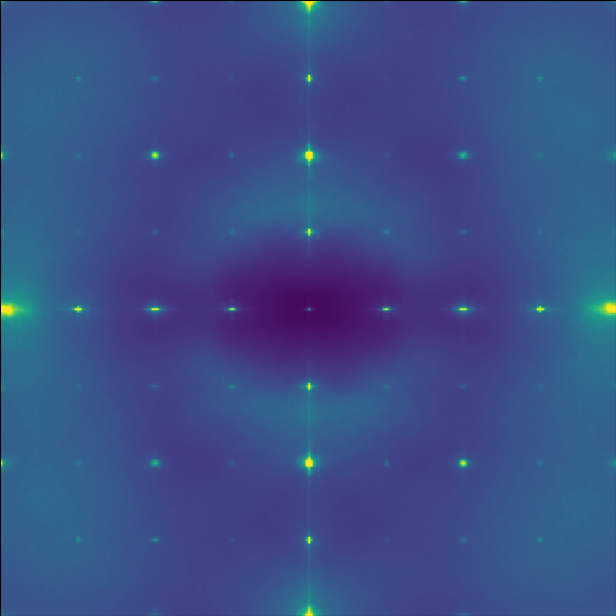} &
        \includegraphics[width=0.15\textwidth]{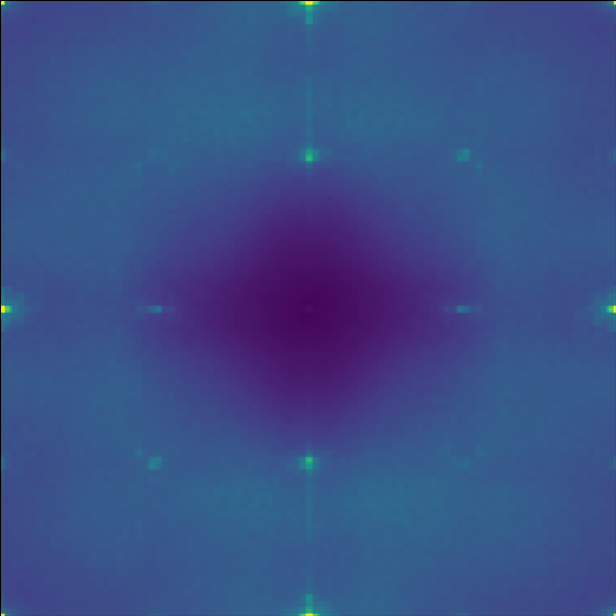} &
        \includegraphics[width=0.15\textwidth]{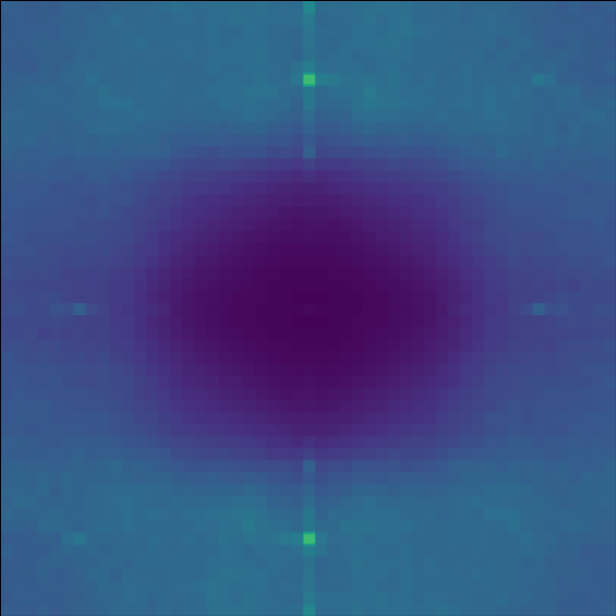} &
        \includegraphics[width=0.15\textwidth]{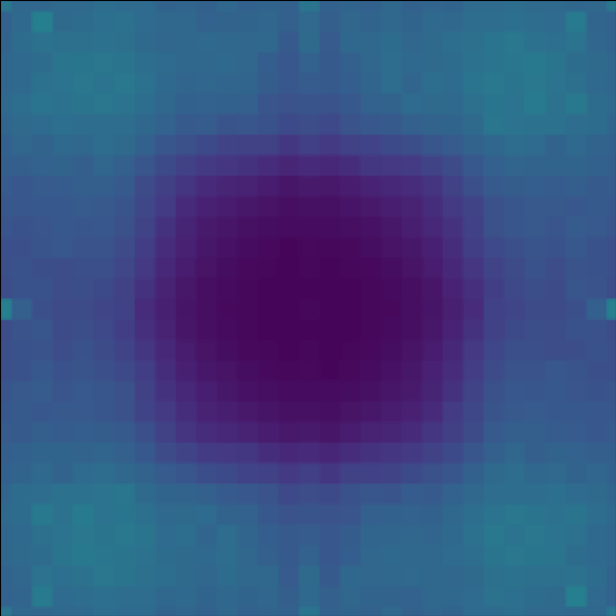} &
        \includegraphics[width=0.15\textwidth]{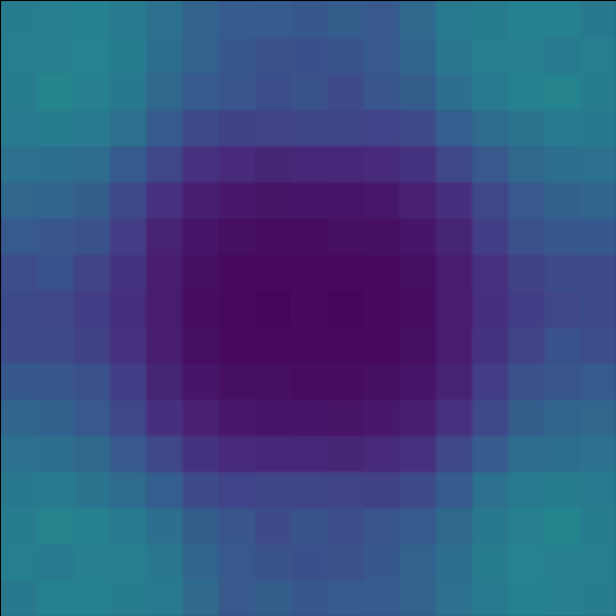} \\
        
        Color Contrast &
        \includegraphics[width=0.15\textwidth]{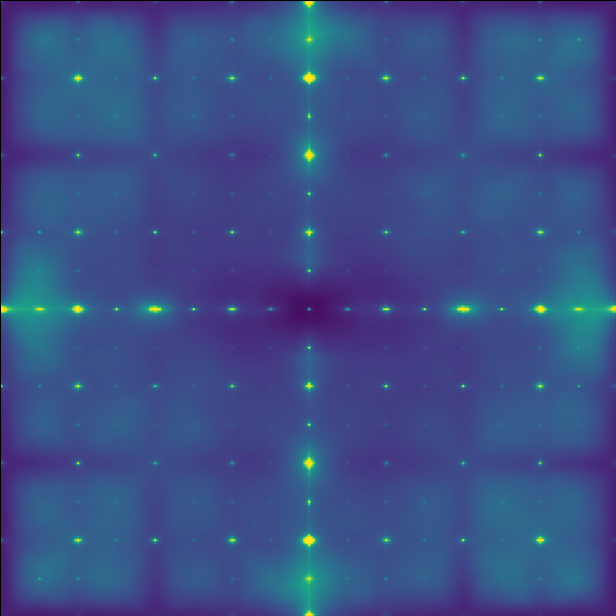} &
        \includegraphics[width=0.15\textwidth]{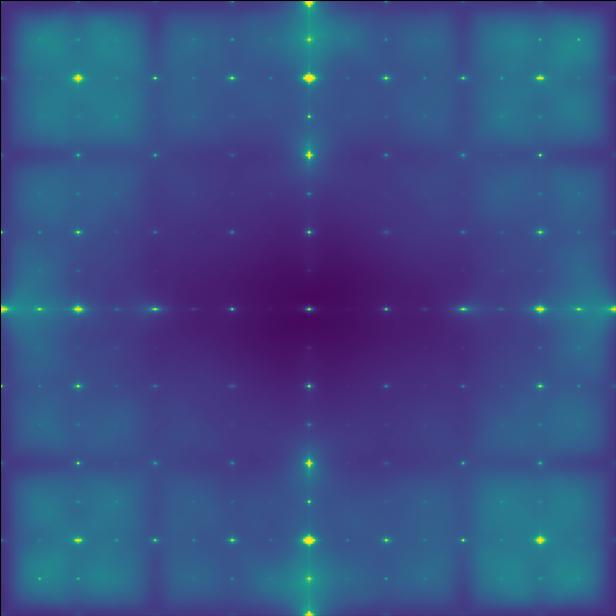} &
        \includegraphics[width=0.15\textwidth]{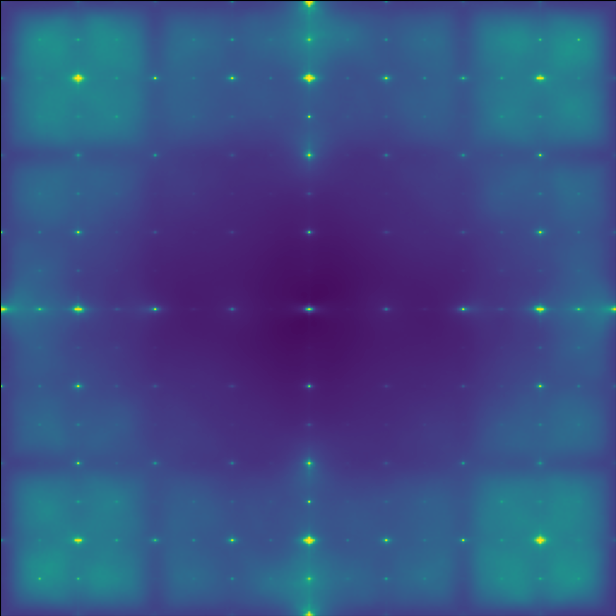} &
        \includegraphics[width=0.15\textwidth]{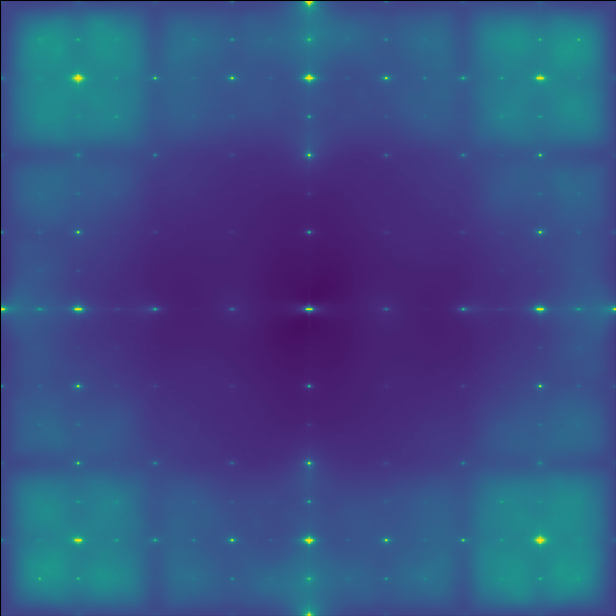} &
        \includegraphics[width=0.15\textwidth]{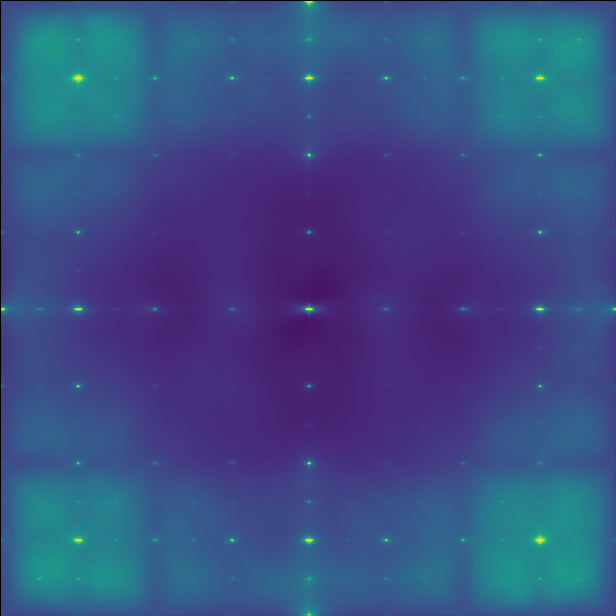} \\
        
        Color Saturation &
        \includegraphics[width=0.15\textwidth]{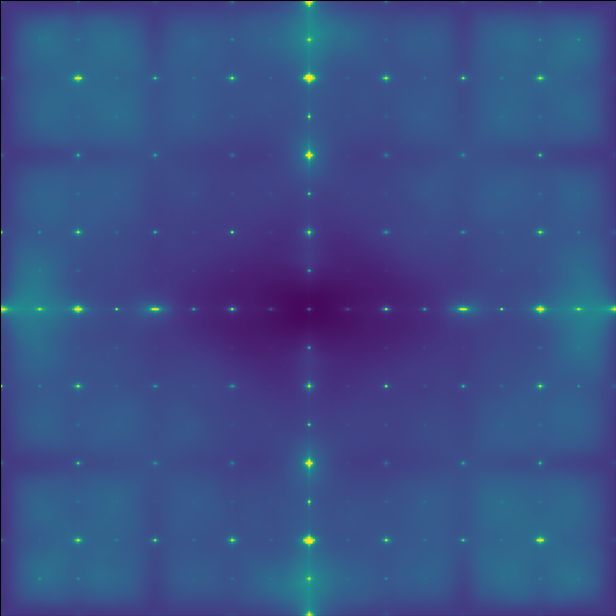} &
        \includegraphics[width=0.15\textwidth]{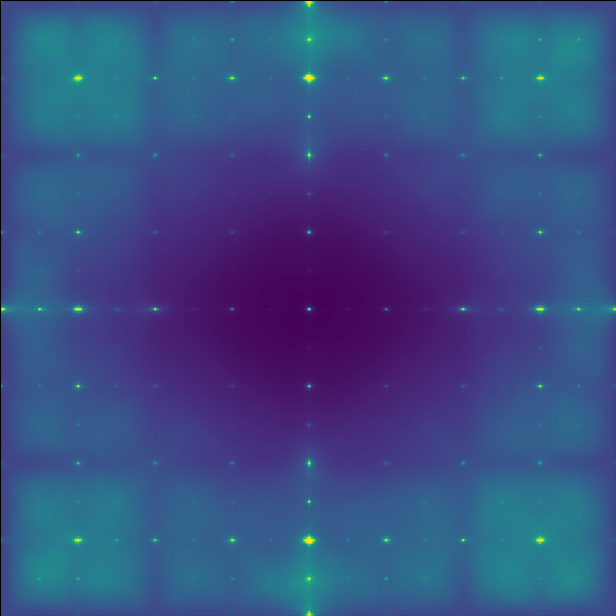} &
        \includegraphics[width=0.15\textwidth]{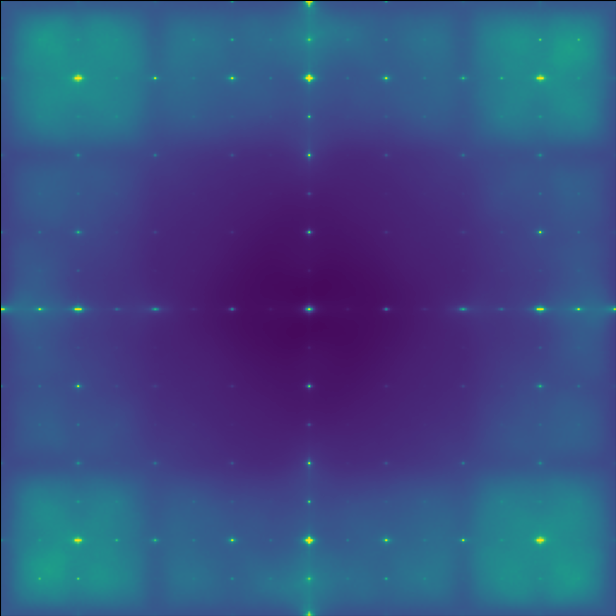} &
        \includegraphics[width=0.15\textwidth]{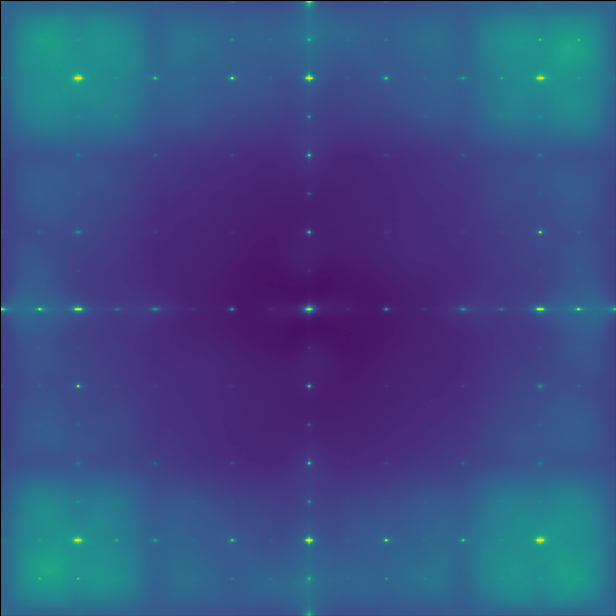} &
        \includegraphics[width=0.15\textwidth]{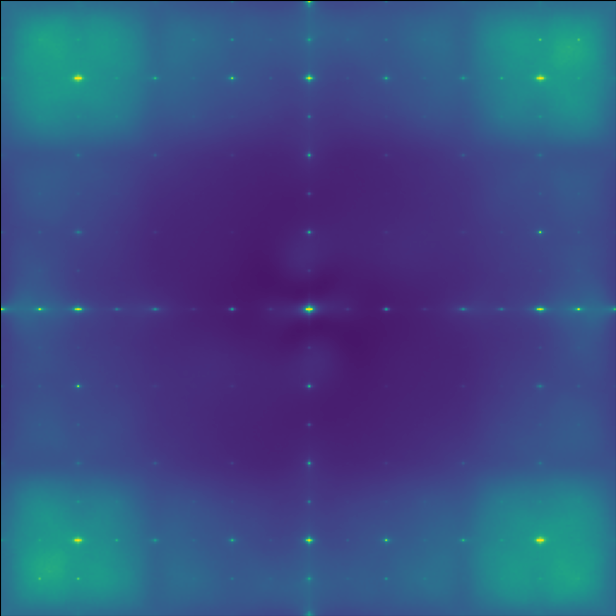}
    \end{tblr}
    \caption{Visualization of frequency domain artifacts in DF images generated by StarGANv2 under varying postprocessing operations. The heatmaps illustrate the spectral signatures across five intensity levels for four different postprocessing techniques: Gaussian blur, resize, color contrast, and color saturation.
    }
    \label{fig:more-spectral-comparison}
\end{figure*}

\section{Proofs} \label{sec:proofs}
\subsubsection{Proof for Theorem \ref{theorem:norm-entropy}}

\begin{proof}

Given a binary classification model $f: \mathcal{X} \rightarrow \mathbb{R}^2$ that produces a probability prediction $0 < p = p(y=1|x) < 1$ for a sample $x$, with $1 - p = p(y=0|x)$ representing the prediction probability of the other class. Let $\hat{y}$ denote the pseudo-label as defined in Eq. \ref{eq:pseudo-label}. We begin by defining two key quantities:

\begin{definition}
    The entropy of prediction in the binary classification is defined as:
    \begin{equation}
        H(x) = -\hat{y}(x) * \log(p) - (-1 -\hat{y}) * \log(1-p)
    \end{equation}
\end{definition}
    
\begin{definition}
    The normalized cross-entropy loss is defined as:
    \begin{equation}
        NCE(x) = \frac{H(x)}{-\log(p) - \log(1-p)}
    \end{equation}
\end{definition}

Let $c = -\log(p) - \log(1-p)$ and suppose that $c$ is a positive constant. We can establish the following equivalence:
\begin{equation}
    H(x) = c*NCE(x)
\end{equation}

Partial derivative of $H(x)$ with respect to $x_i, \; i = 1, \dots, n$:
\begin{align}
    \frac{\partial H(x)}{\partial x_i} &= \frac{\partial}{\partial x_i}[-\hat{y}(x)\log(p) - (1-\hat{y})\log(1-p)] \\
    &= -\frac{\partial \hat{y}(x)}{\partial x_i}\log(p) + \frac{\partial \hat{y}(x)}{\partial x_i}\log(1-p) \\
    &= -\frac{\partial \hat{y}(x)}{\partial x_i}[\log(p) - \log(1-p)] \\
    &= -\frac{\partial \hat{y}(x)}{\partial x_i}\log\left(\frac{p}{1-p}\right)
\end{align}

Partial derivative of $NCE(x)$ with respect to $x_i , \; i = 1, \dots, n$:
\begin{align}
    \frac{\partial NCE(x)}{\partial x_i} &= \frac{\partial}{\partial x_i}\left(\frac{H(x)}{c}\right) \\
    &= \frac{1}{c} \cdot \frac{\partial H(x)}{\partial x_i} \\
    &= -\frac{1}{c} \cdot \frac{\partial \hat{y}(x)}{\partial x_i}\log\left(\frac{p}{1-p}\right)
\end{align}

We have:
\begin{align}
    \frac{\partial H(x)}{\partial x_i} = 0 &\iff -\frac{\partial \hat{y}(x)}{\partial x_i}\log\left(\frac{p}{1-p}\right) = 0 \\
    &\iff c \cdot \left(-\frac{1}{c} \cdot \frac{\partial \hat{y}(x)}{\partial x_i}\log\left(\frac{p}{1-p}\right)\right) = 0 \\
    &\iff c \cdot \frac{\partial NCE(x)}{\partial x_i} = 0 \\
    &\iff \frac{\partial NCE(x)}{\partial x_i} = 0
\end{align}

The last equivalence holds because $c$ is positive. Therefore, for all $i = 1,\ldots,n$:
\begin{equation}
    \frac{\partial H(x)}{\partial x_i} = 0 \iff \frac{\partial NCE(x)}{\partial x_i} = 0
\end{equation}

This equivalence proves that the partial derivatives of both $H(x)$ and $NCE(x)$ vanish at the same points. Since $c$ is positive, $H(x)$ has a local extremum if and only if $NCE(x)$ has a local extremum at the same point $x^*$.

\end{proof}

\subsubsection{Proof for Lemma \ref{lemma:conv-frequency}}

We need Definition \ref{def:dct} to prove the Lemma \ref{lemma:conv-frequency}.


Note that, for simplification, in this proof, we assume that $x_1$ and $x_2$ have the same size $M \times N$.
\begin{proof}
    The spatial domain convolution of two images is given by:
    \begin{equation}
    (x_1 \circledast x_2)(m.n) =  \frac{1}{MN} \sum^{M-1}_{k=0} \sum^{N-1}_{l=0} x_1(k,l)x_2(m-k,n-l)
    \label{eq:2}
    \end{equation}
    Take the Fourier transform of Eq. \ref{eq:2}, we obtain:
    \begin{equation}
        \begin{split}
            &\mathcal{F}\{x_1 \otimes x_2\} = \\ 
            &\sum_{m=0}^{M-1} \sum_{n=0}^{N-1} 
            \left[\sum_{k=0}^{M-1} \sum_{l=0}^{N-1} x_1(k,l)x_2(m-k,n-l)\right] 
            e^{-j2\pi(\frac{um}{M} + \frac{vn}{N})} \\
            &\sum_{k=0}^{M-1} \sum_{l=0}^{N-1} x_1(k,l)
            \left[\sum_{m=0}^{M-1} \sum_{n=0}^{N-1} x_2(m-k,n-l)e^{-j2\pi(\frac{um}{M} + \frac{vn}{N})}\right]
        \end{split}
    \end{equation}
    After change of variables $p = m-k$, $q = n-l$ and substitution:
    \begin{equation}
        \begin{split}
            &\mathcal{F}\{x_1 \otimes x_2\} = \\
            &\sum_{k=0}^{M-1} \sum_{l=0}^{N-1} x_1(k,l)
            \left[\sum_{p=0}^{M-1} \sum_{q=0}^{N-1} x_2(p,q)e^{-j2\pi(\frac{u(p+k)}{M} + \frac{v(q+l)}{N})}\right] \\
            &= \sum_{k=0}^{M-1} \sum_{l=0}^{N-1} x_1(k,l)e^{-j2\pi(\frac{uk}{M} + \frac{vl}{N})} \\
            &\cdot \left[\sum_{p=0}^{M-1} \sum_{q=0}^{N-1} x_2(p,q)e^{-j2\pi(\frac{up}{M} + \frac{vq}{N})}\right]
        \end{split}
    \end{equation}    

    By definition of the 2D-DFT \ref{def:dct}, we can recognize:
    \begin{equation}
        \begin{cases}
            \sum_{k=0}^{M-1} \sum_{l=0}^{N-1} x_1(k,l)e^{-j2\pi(\frac{uk}{M} + \frac{vl}{N})} = X_1(u,v) \\
            \sum_{p=0}^{M-1} \sum_{q=0}^{N-1} x_2(p,q)e^{-j2\pi(\frac{up}{M} + \frac{vq}{N})} = X_2(u,v)
        \end{cases}
    \end{equation}
    Therefore:
    \begin{equation}
        \mathcal{F}\{x_1 \otimes x_2\} = X_1(u,v) \cdot X_2(u,v)
    \end{equation}
    
\end{proof}

\section{More Experiments of Generation Artifacts}\label{sec:more-artifacts}
Figure \ref{fig:more-spectral-comparison} illustrates frequency spectra of the fake sample generated by StarGAN2. This fake sample is applied by 4 types of postprocessing operations, with the intensity level increasing from $1$ to $5$. Figure \ref{fig:exp_cross_augs} shows the performance degradation of DF detectors under different types of postprocessing techniques across 5 intensity levels. Note that, in this experimental evaluation, both training and test samples are drawn from the same underlying data distribution (FaceForensics++ \cite{rossler2019faceforensics++}), and only postprocessing operations are unseen during the testing phase.

\section{Experimental Details} \label{sec:experiment-details}
\subsection{Datasets}
\subsubsection{Training dataset.} We use FF++ \cite{rossler2019faceforensics++} for training the source model (Xception) and other DF detectors. In this dataset, real videos collected from YouTube, which are then used to generate fake videos through four DF methods, including DeepFake, Face2Face, FaceSwap, and NeuralTexture. FF++ contains a total of 5000 videos, of which 1000 videos are sourced from YouTube. 

\subsubsection{Test datasets.} 
To evaluate the adaptability of our \texttt{T$^2$A} method, we use six datasets at the inference time, including:
\begin{itemize}
    \item CelebDF-v1 and CelebDF-v2 \cite{li2020celeb}: contain 998 real videos collected from 59 celebrities and 6434 fake videos improved by using techniques, such as higher resolution synthesis, color mismatch reduction, improved face mask, temporal flickering reduction. Videos in CelebDF datasets are variations in face sizes, orientations, lighting condiitons and backgrounds.
    \item DeepFakeDetection (DFD) \cite{DFD}: includes 363 real videos and 3000 fake videos.
    \item DeepFake Detection Challenge Preview (DFDCP) \cite{dolhansky2019dee}: consists of 1131 real videos of 66 individuals total and 4119 fake videos generated by multiple synthesis methods. Videos include varied lighting conditions, head poses, and backgrounds.
    \item UADFV \cite{li2018ictu} which is composed of 98 real and fake videos from 49 different identities. This dataset mainly focuses on blinking, assisting in DF detection through physiological signals.
    \item FaceShifter (FSh) \cite{li2020advancing}: includes a total of 2000 real and fake videos.
\end{itemize}

The image size of training and test samples are $256 \times 256$ unless using Resize postprocessing (described in Sec. \ref{ssec:postprocessing-intensity}). During the testing phase, individual frames extracted from these videos serve as our evaluation data. We use test sets of these datasets provided by \cite{yan2023deepfakebench}. 

\subsection{Intensity Levels of Postprocessing Techniques}
\label{ssec:postprocessing-intensity}
\begin{figure*}[!ht]
    \centering
    \begin{subfigure}[t]{0.95\linewidth}
        \includegraphics[width=0.95\linewidth]{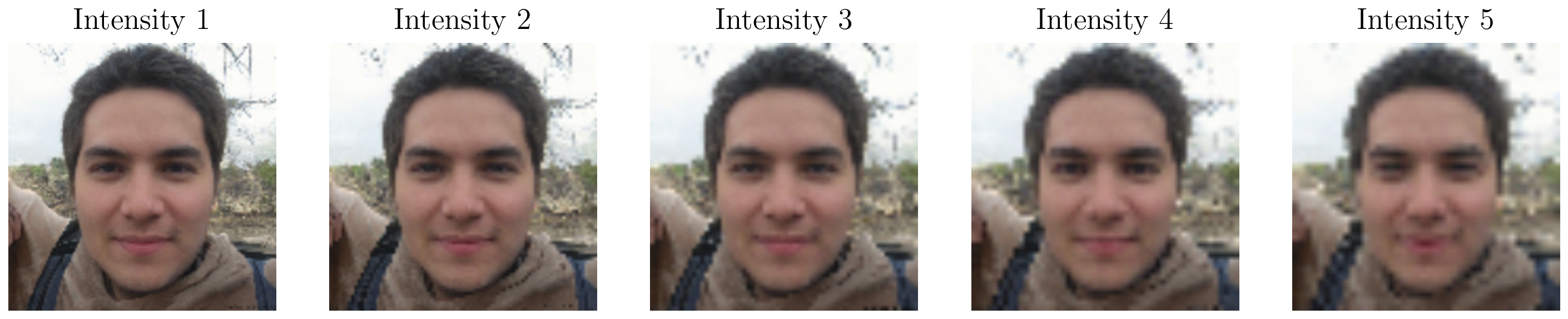}
        \caption{Resize}
    \end{subfigure}
    \\
    \begin{subfigure}[t]{0.95\linewidth}
        \includegraphics[width=0.95\linewidth]{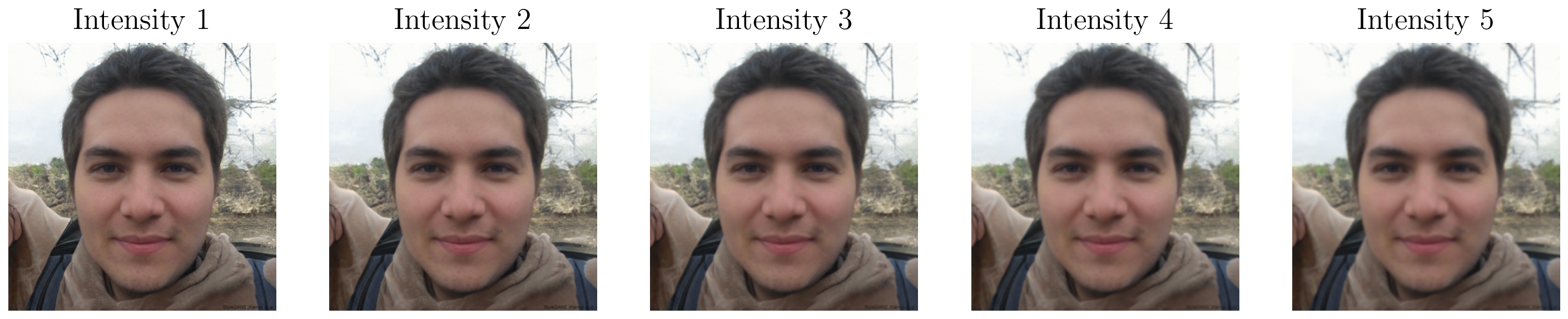}
        \caption{Gaussian Blur}
    \end{subfigure}
    \begin{subfigure}[t]{0.95\linewidth}
        \includegraphics[width=0.95\linewidth]{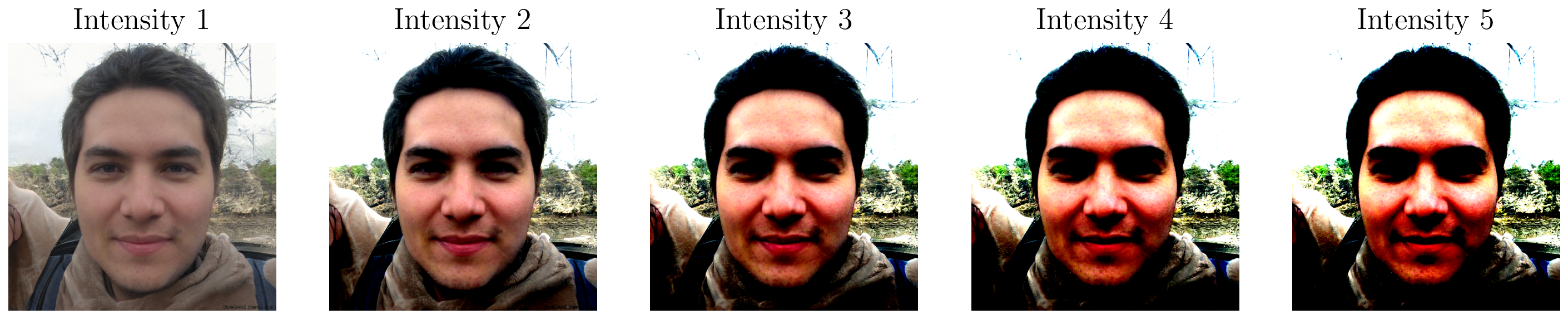}
        \caption{Color Contrast}
    \end{subfigure}
    \\
    \begin{subfigure}[t]{0.95\linewidth}
        \includegraphics[width=0.95\linewidth]{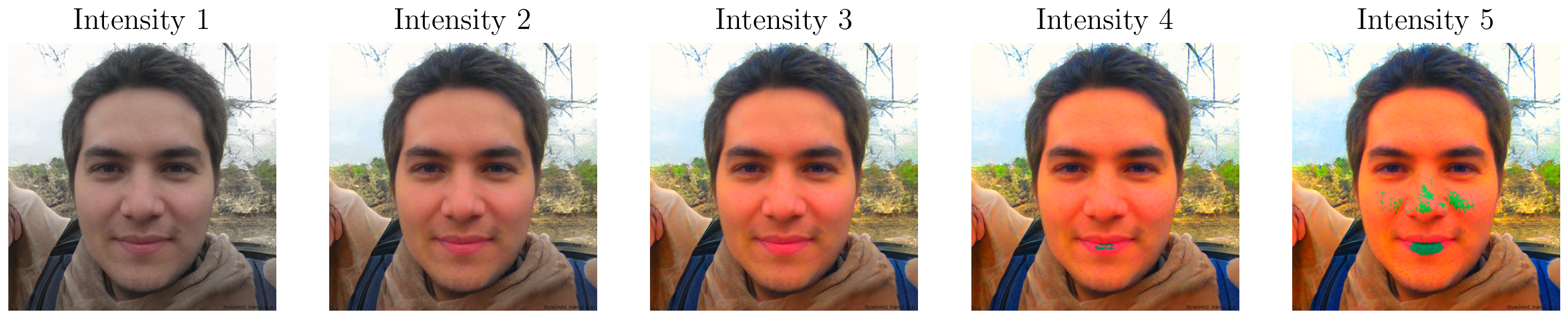}
        \caption{Color Saturation}
    \end{subfigure}
    \caption{Four postprocessing operation types across five intensity levels.}
    \label{fig:intensity-level}
\end{figure*}

In practice, both authentic and manipulated images frequently undergo various postprocessing operations. For real-world DF detection requirements, resilience to unknown postprocessing techniques is crucial. Following \cite{chen2022ost}, we evaluate detector robustness across four fundamental postprocessing operations: Gaussian blur, resize, color saturation, and color contrast. For each operation, we implement five intensity levels based on standard corruption benchmarking practices \cite{hendrycks2019benchmarking}. Figure \ref{fig:intensity-level} shows an example of the five intensity levels of four types of postprocessing techniques.

Regarding Gaussian blur operation, we employ progressively larger kernel sizes: $5 \times 5$, $9 \times 9$, $13 \times 13$, $17 \times 17$, and $21 \times 21$ (levels $1-5$, respectively). Each larger kernel size produces a progressively stronger blurring effect on the image. For the resize operation, we first downsample to a smaller resolution and then upsample back to $256 \times 256$, creating progressively stronger image quality degradation as more pixel information is lost at lower intermediate resolutions. For each intensity level, the intermediate resolution is: $128$, $85$, $64$, $51$, and $41$, respectively. 

To manipulate color saturation across 5 intensity levels, we convert the image from BGR to YCbCr color space, where Y represents luminance and Cb/Cr represents chrominance components, then a saturation factor $i$ (each intensity level) is applied to linearly push Cb and Cr values $c$ away from the center point ($128$) while preserving Y (luminance) by the transformation as follows: $c:= 128 + (c - 128) * i$.

For color contrast operation, we modify image contrast across 5 intensity levels by manipulating pixel values around their mean while applying channel-wise enhancements. In particular, for intensity $i$, the pixel value $c$ will be updated as follows: $c := \mathbb{E}_{c} + (c-\mathbb{E}_{c})i$. Then, the pixel values will be clipped to the range of $[0, 255]$ to preserve brightness.



\label{sssection:analysis-losses}
\begin{figure*}[ht]
    \centering
    \begin{subfigure}[t]{0.3\linewidth}
        \includegraphics[width=0.95\linewidth]{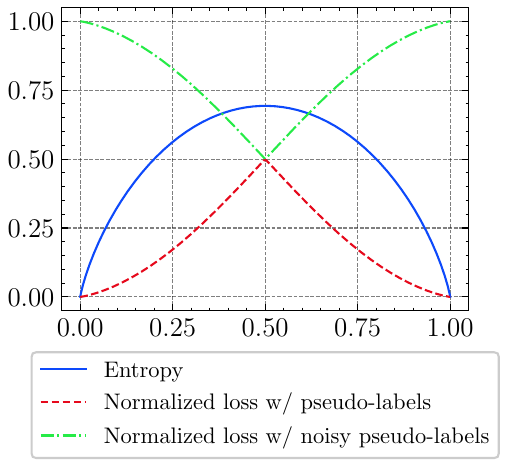}
        \caption{Normalized loss with pseudo label $\mathcal{L}_{norm}(x, \hat{y})$ and noisy pseudo-label $\mathcal{L}_{nn}(x,\tilde{y})$. $\mathcal{L}_{nn}(x,\tilde{y})$ is the opposite of $\mathcal{L}_{norm}(x, \hat{y})$.}
    \end{subfigure}
    \hfill
    \begin{subfigure}[t]{0.3\linewidth}
        \includegraphics[width=0.95\linewidth]{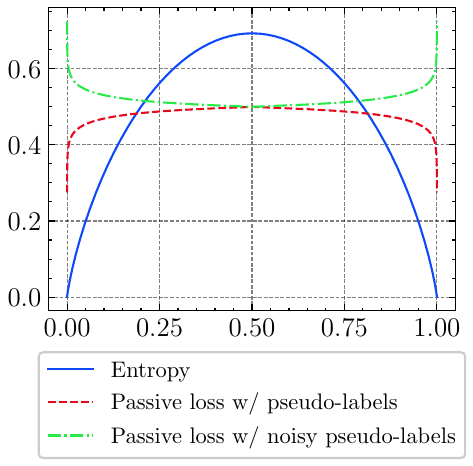}
        \caption{Passive loss function with pseudo label $\mathcal{L}_{p}(x, \hat{y})$ and noisy pseudo-label $\mathcal{L}_{p}(x,\tilde{y})$. $\mathcal{L}_{p}(x,\tilde{y})$ is the opposite of $\mathcal{L}_{p}(x, \hat{y})$.}
    \end{subfigure}
    \hfill
    \begin{subfigure}[t]{0.3\linewidth}
        \includegraphics[width=0.95\linewidth]{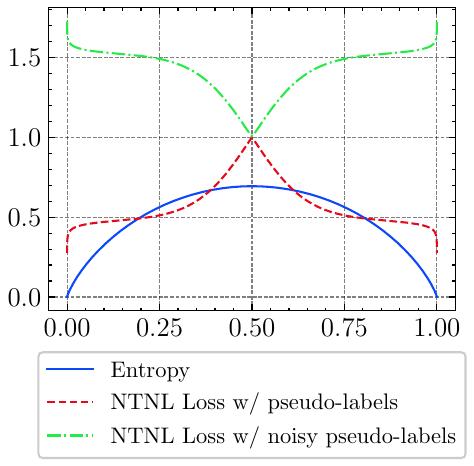}
        \caption{Noise-tolerant negative loss (NTNL) functions with pseudo label $\mathcal{L}_{NTNL}(x, \hat{y})$ and noisy-pseudo label $\mathcal{L}_{NTNL}(x,\tilde{y})$. $\mathcal{L}_{NTNL}(x,\tilde{y})$ is the opposite of $\mathcal{L}_{NTNL}(x, \hat{y})$.}
    \end{subfigure}
    \caption{Comparison of different loss functions against entropy minimization. Each plot demonstrates how the proposed loss functions exhibit complementary behavior to entropy minimization across different prediction probabilities.}
    \label{fig:proposed-loss}
\end{figure*}

\subsection{Implementation Details}
\subsubsection{TTA baselines}
For all TTA approaches, TENT \cite{wang2020tent}, MEMO \cite{zhang2022memo}, EATA \cite{niu2022efficient}, CoTTA \cite{wang2022continual}, LAME \cite{boudiaf2022parameter}, ViDA \cite{liuvida2023}, and COME \cite{zhang2024come}, we follow all hyperparameters that are set in their Github unless it does not provide.

\subsubsection{DF Detection baselines}
Since the pre-trained models of EfficientNetB4 \cite{tan2019efficientnet}, F3Net \cite{qian2020thinking}, CORE \cite{ni2022core}, and RECCE \cite{cao2022end} are not provided, we use that provided by \cite{yan2023deepfakebench}. 

\subsubsection{Hyperparameters}
Table \ref{tab:hyperparams} provides hyperparameters details.

\begin{table}[ht]
\centering
\caption{Hyperparameters.}
\label{tab:hyperparams}
\begin{tblr}{
  width = \linewidth,
  colspec = {Q[473]Q[406]},
  cells = {c},
  vline{2} = {-}{},
  hline{2} = {-}{},
}
\textbf{Hyperparameter}   & \textbf{Values}      \\
$\alpha$ & $\{1.0, 2.0\}$ \\
$\beta$  & $\{1.0, 2.0\}$ \\
$\psi$   & $\{0.01, 0.1\}$     
\end{tblr}
\end{table}

\begin{table}[ht]
\centering
\fontsize{6pt}{6pt}\selectfont
\caption{Effectiveness of components in \texttt{T$^2$A} method on FF++ dataset. The results are averaged across 4 postprocessing techniques with 5 intensity levels.}
\label{tab:ablation}
\begin{tblr}{
  width = \linewidth,
  colspec = {Q[30]Q[40]Q[40]Q[40]Q[50]Q[30]Q[30]Q[30]},
  cells = {c},
  vline{2,6} = {-}{},
  hline{1,7} = {-}{0.08em},
  hline{2} = {-}{},
  hline{2} = {2}{-}{},
}
             & \textbf{Using} $\mathcal{L}_{EM}$ & \textbf{Using} $\mathcal{L}_{nn}$ & \textbf{Using} $\mathcal{L}_{p}$ & \textbf{Gradients masking} & \textbf{ACC}        & \textbf{AUC}        & \textbf{AP}         \\
\texttt{T$^2$A} & \ding{51}                              &                                        &                                       &                            & 0.8491 $\pm$ 0.01 & 0.8542 $\pm$ 0.02 & 0.9570 $\pm$ 0.01 \\
\texttt{T$^2$A} & \ding{51}                              & \ding{51}                              &                                       &                            & 0.8472 $\pm$ 0.01 & 0.8580 $\pm$ 0.02 & 0.9583 $\pm$ 0.01 \\
\texttt{T$^2$A} & \ding{51}                              &                                        & \ding{51}                             &                            & 0.8490 $\pm$ 0.01 & 0.8542 $\pm$ 0.02 & 0.9570 $\pm$ 0.01 \\
\texttt{T$^2$A} & \ding{51}                              & \ding{51}                              & \ding{51}                             &                            & 0.8394 $\pm$ 0.01 & 0.8646 $\pm$ 0.01 & 0.9617 $\pm$ 0.01 \\
\texttt{T$^2$A} & \ding{51}                              & \ding{51}                              & \ding{51}                             & \ding{51}                  & 0.8582 $\pm$ 0.01 & 0.8813 $\pm$ 0.01 & 0.9664 $\pm$ 0.01 
\end{tblr}
\end{table}

\begin{figure}[ht]
    \centering
    \includegraphics[width=0.45\textwidth]{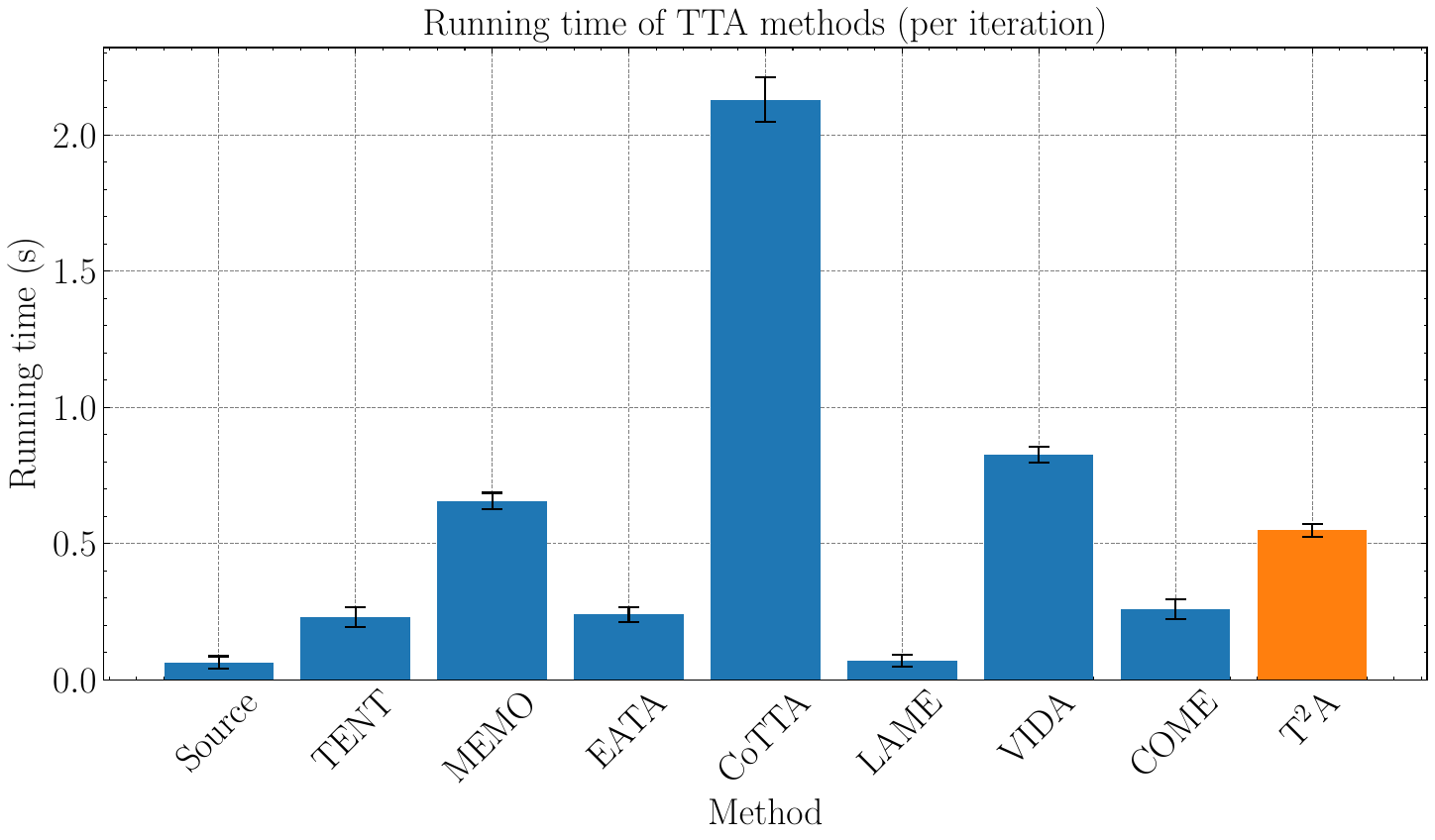}
    \caption{Average running time per iteration of TTA methods.}
    \label{fig:running-time}
\end{figure}

\section{More Experimental Results}
\label{sec:more-results}
\subsection{Ablation Study} \label{subsec:ablation-study}

\subsubsection{Analysis on Components of \texttt{T$^2$A} method}
Our method consists of three main components: 1) Entropy Minimization (EM) loss, 2) Noise-tolerant negative loss (NTNL), and 3) Gradients masking. We ablate them in Table \ref{tab:ablation}. Compared with the EM loss, our proposed method (5-th row) achieves better performance across three metrics. This validates our motivation that some overconfident samples (i.e., optimized by EM) hurt the performance of the model during adaptation. We also evaluate the impact of the normalized negative loss $\mathcal{L}_{nn}$ and the passive loss $\mathcal{L}_p$ on the adaptation performance of the model.

\subsubsection{Analysis on Proposed Loss Functions}
We provide an analysis of our proposed loss functions in comparison with EM, demonstrating their complementary behavior. Figure \ref{fig:proposed-loss} illustrates three variants of our negative learning approach and their relationships with EM across the probability range $[0,1]$. The figure shows that three variants of loss functions with noisy pseudo-labels (green line) exhibit opposing behavior with the EM (blue line). 

\subsection{Full results of comparison with SoTA TTA methods under unknown postprocessing techniques}
In Table \ref{tab:TTA-post-process-appendix}, we provide more results to compare our \texttt{T$^2$A} method with SoTA TTA approaches on FF++ with the intensity level from 1 to 5. Although the source model achieves better performance at the lowest intensity level (level 1) for color contrast and color saturation operations, our method exhibits consistently better adaptation performance as the intensity of postprocessing increases. Across all four postprocessing types, T2A generally outperforms existing TTA approaches across all three evaluation metrics, demonstrating particular resilience to more aggressive postprocessing manipulations.

\subsection{Full results of improvements of DF detectors under unknown postprocessing techniques}
Table \ref{tab:DF-post-process-appendix} shows more results about improvement of DF detectors under unknown postprocessing techniques scenario with the intensity level 1-5. THe table shows that our method can improve the adaptation performance of DF detectors across intensity levels of postprocessing techniques.

\subsection{Wall-clock running time of T$^2$A} \label{subsec:running-time}
We report the running time per iteration of TTA methods. Figure \ref{fig:running-time} compares the running time of our method and other TTA approaches. Experiments on the DFDCP dataset, performed using an NVIDIA RTX 4090 GPU. Our \texttt{T$^2$A} method achieves superior adaptation performance ($73.2\%$ AUC) within $0.5s$ per iteration. EATA, COME, and TENT demonstrate comparable execution times (approximately $0.26s$) but with lower performance ($70.04\%$, $70.13\%$, and $69.9\%$ AUC, respectively). While LAME achieves the fastest execution ($0.07s$), it shows significantly degraded performance ($59.88\%$ AUC). Conversely, methods employing extensive augmentation during adaptation—MEMO ($0.66s$), VIDA ($0.83s$), and CoTTA ($2.23s$) — incur substantially higher computational costs. Note that, TENT, EATA, COME, LAME achieve running efficiency due to the adaptation being applied to BN layers only. This shows that our method achieves an effective balance between computational efficiency and adaptation performance.

\begin{table*}[ht]
\centering
\fontsize{6pt}{6pt}\selectfont
\caption{Comparison with state-of-the-art TTA methods on FF++ with different postprocessing techniques from intensity level from 1 to 5. The bold number indicates the best result.}
\label{tab:TTA-post-process-appendix}
\begin{tblr}{
  width = \linewidth,
  colspec = {Q[70]Q[35]Q[35]Q[35]Q[35]Q[35]Q[35]Q[35]Q[35]Q[35]Q[35]Q[35]Q[35]Q[35]Q[35]Q[35]},
  cells = {c},
  cell{1}{1} = {r=2}{},
  cell{1}{2} = {c=3}{0.1\linewidth},
  cell{1}{5} = {c=3}{0.1\linewidth},
  cell{1}{8} = {c=3}{0.1\linewidth},
  cell{1}{11} = {c=3}{0.1\linewidth},
  cell{1}{14} = {c=3}{0.1\linewidth},
  cell{3}{1} = {c=16}{0.1\linewidth},
  cell{13}{1} = {c=16}{0.1\linewidth},
  cell{23}{1} = {c=16}{0.1\linewidth},
  cell{33}{1} = {c=16}{0.1\linewidth},
  cell{43}{1} = {c=16}{0.1\linewidth},
  vline{2,5,8,11,14} = {1-3, 4-12,14-22,24-32,34-42,44-52}{},
  hline{1,53} = {-}{0.08em},
  hline{2} = {2-16}{},
  hline{3-4,13-14,23-24,33-34,43-44} = {-}{},
  hline{3-4,13-14,23-24,33-34,43-44} = {2}{-}{},
}
\textbf{Method }              & \textbf{Color Contrast } &                 &                 & \textbf{Color Saturation } &                 &                 & \textbf{Resize } &                 &                 & \textbf{Gaussian Blur } &                 &                 & \textbf{Average } &                 &                 \\
                              & \textbf{ACC}             & \textbf{AUC}    & \textbf{AP}     & \textbf{ACC}               & \textbf{AUC}    & \textbf{AP}     & \textbf{ACC}     & \textbf{AUC}    & \textbf{AP}     & \textbf{ACC}            & \textbf{AUC}    & \textbf{AP}     & \textbf{ACC}      & \textbf{AUC}    & \textbf{AP}     \\
\textbf{Intensity level = 1 } &                          &                 &                 &                            &                 &                 &                  &                 &                 &                         &                 &                 &                   &                 &                 \\
Source                        & \textbf{0.9171}          & \textbf{0.9604} & \textbf{0.9902} & \textbf{0.9214}            & \textbf{0.9602} & \textbf{0.9901} & 0.9042           & 0.9469          & 0.9867          & 0.9028                  & 0.9481          & 0.9869          & \textbf{0.9114}   & 0.9539          & 0.9884          \\
TENT                          & 0.9100                   & 0.9556          & 0.9887          & 0.8914                     & 0.9468          & 0.9859          & 0.9042           & 0.9456          & 0.9828          & 0.9042                  & 0.9488          & 0.9868          & 0.9024            & 0.9492          & 0.9860          \\
MEMO                          & 0.8657                   & 0.9307          & 0.9824          & 0.8657                     & 0.9284          & 0.9815          & 0.8585           & 0.9285          & 0.9812          & 0.8557                  & 0.9281          & 0.9814          & 0.8614            & 0.9289          & 0.9816          \\
EATA                          & 0.9100                   & 0.9558          & 0.9887          & 0.9085                     & 0.9550          & 0.9888          & 0.9071           & 0.9415          & 0.9827          & 0.9042                  & 0.9489          & 0.9868          & 0.9074            & 0.9503          & 0.9867          \\
CoTTA                         & 0.8928                   & 0.9447          & 0.9863          & 0.8942                     & 0.9437          & 0.9858          & 0.8885           & 0.9310          & 0.9823          & 0.8885                  & 0.9333          & 0.9830          & 0.8910            & 0.9381          & 0.9843          \\
LAME                          & 0.8171                   & 0.9134          & 0.9668          & 0.8185                     & 0.9164          & 0.9684          & 0.8071           & 0.8874          & 0.9557          & 0.8157                  & 0.9012          & 0.9624          & 0.8146            & 0.9046          & 0.9633          \\
VIDA                          & 0.8771                   & 0.9330          & 0.9827          & 0.8771                     & 0.9324          & 0.9826          & 0.8828           & 0.9309          & 0.9823          & 0.8785                  & 0.9315          & 0.9824          & 0.8789            & 0.9319          & 0.9825          \\
COME                          & 0.9000                   & 0.9536          & 0.9885          & 0.9071                     & 0.9524          & 0.9882          & 0.8985           & 0.9453          & 0.9862          & 0.8971                  & 0.9455          & 0.9863          & 0.9007            & 0.9492          & 0.9873          \\
\texttt{T$^2$A} (Ours)                    & 0.9128                   & 0.9562          & 0.9888          & 0.9100                     & 0.9559          & 0.9888          & \textbf{0.9071}  & \textbf{0.9485} & \textbf{0.9867} & \textbf{0.9071}         & \textbf{0.9588} & \textbf{0.9888} & 0.9092            & \textbf{0.9549} & \textbf{0.9882} \\
\textbf{Intensity level = 2}  &                          &                 &                 &                            &                 &                 &                  &                 &                 &                         &                 &                 &                   &                 &                 \\
Source                        & 0.8600                   & 0.9233          & 0.9794          & 0.8900                     & 0.9381          & 0.9840          & 0.8671           & 0.9165          & 0.9789          & 0.8757                  & 0.9169          & 0.9788          & 0.8732            & 0.9217          & 0.9803          \\
TENT                          & 0.8928                   & 0.9150          & 0.9700          & 0.9000                     & 0.9453          & 0.9861          & 0.8728           & 0.9206          & 0.9795          & 0.8842                  & 0.9241          & 0.9807          & 0.8873            & 0.9267          & 0.9775          \\
MEMO                          & 0.8342                   & ~0.8791         & 0.9662          & 0.8342                     & 0.9021          & 0.9744          & 0.8400           & 0.8898          & 0.9706          & 0.8442                  & 0.9019          & 0.9741          & 0.8382            & 0.8932          & 0.9713          \\
EATA                          & 0.8928                   & 0.9151          & 0.9800          & 0.9000                     & 0.9451          & 0.9861          & 0.8700           & 0.9208          & 0.9795          & 0.8842                  & 0.9243          & 0.9808          & 0.8867            & 0.9274          & 0.9806          \\
CoTTA                         & ~0.8785                  & 0.9046          & 0.9749          & 0.8900                     & 0.9359          & 0.9835          & 0.8714           & 0.9023          & 0.9748          & 0.8742                  & 0.8996          & 0.9736          & 0.8785            & 0.9106          & 0.9767          \\
LAME                          & 0.8457                   & 0.8873          & 0.9598          & 0.8128                     & 0.8943          & 0.9603          & 0.8057           & 0.8113          & 0.9219          & 0.8042                  & 0.8582          & 0.9457          & 0.8171            & 0.8628          & 0.9469          \\
VIDA                          & 0.8514                   & 0.8912          & 0.9704          & 0.8585                     & 0.9241          & 0.9801          & 0.8471           & 0.9055          & 0.9757          & 0.8685                  & 0.9082          & 0.9757          & 0.8564            & 0.9072          & 0.9753          \\
COME                          & 0.8785                   & 0.9108          & 0.9692          & 0.8900                     & 0.9442          & \textbf{0.9862} & 0.8785           & 0.9173          & 0.9786          & 0.8828                  & 0.9198          & 0.9791          & 0.8825            & 0.9235          & 0.9787          \\
\texttt{T$^2$A} (Ours)                    & \textbf{0.8885}          & \textbf{0.9251} & \textbf{0.9800} & \textbf{0.9042}            & \textbf{0.9456} & \textbf{0.9862} & \textbf{0.8742}  & \textbf{0.9256} & \textbf{0.9798} & \textbf{0.8871}         & \textbf{0.9252} & \textbf{0.9810} & \textbf{0.8882}   & \textbf{0.9321} & \textbf{0.9817} \\
\textbf{Intensity level = 3}  &                          &                 &                 &                            &                 &                 &                  &                 &                 &                         &                 &                 &                   &                 &                 \\
Source                        & 0.7928                   & 0.8683          & 0.9630          & 0.8214                     & 0.8365          & 0.9467          & 0.8357           & 0.8800          & 0.9697          & 0.8271                  & 0.8594          & 0.9609          & 0.8192            & 0.8610          & 0.9600          \\
TENT                          & 0.8728                   & 0.8874          & 0.9723          & 0.8471                     & 0.8652          & 0.9595          & 0.8500           & 0.8869          & 0.9637          & 0.8628                  & 0.8758          & 0.9590          & 0.8581            & 0.8788          & 0.9636          \\
MEMO                          & 0.8200                   & 0.8454          & 0.9572          & 0.8200                     & 0.8476          & 0.9567          & 0.8300           & 0.8775          & 0.9681          & 0.8342                  & 0.8672          & 0.9639          & 0.8260            & 0.8594          & 0.9614          \\
EATA                          & 0.8728                   & 0.8876          & 0.9723          & 0.8457                     & 0.8747          & 0.9594          & 0.8528           & 0.8875          & 0.9639          & 0.8628                  & 0.8763          & 0.9592          & 0.8585            & 0.8815          & 0.9637          \\
CoTTA                         & 0.8514                   & 0.8669          & 0.9628          & 0.8357                     & 0.8552          & 0.9620          & 0.8528           & 0.8683          & 0.9651          & 0.8571                  & 0.8696          & 0.9636          & 0.8492            & 0.8650          & 0.9634          \\
LAME                          & 0.8042                   & 0.8455          & 0.9465          & 0.8042                     & 0.7832          & 0.9199          & 0.8057           & 0.8167          & 0.9300          & 0.8042                  & 0.7452          & 0.8993          & 0.8046            & 0.7977          & 0.9239          \\
VIDA                          & 0.8471                   & 0.8757          & 0.9648          & 0.8285                     & 0.8587          & 0.9630          & 0.8371           & 0.8781          & 0.9654          & 0.8400                  & 0.8627          & 0.9605          & 0.8382            & 0.8688          & 0.9634          \\
COME                          & 0.8571                   & 0.8897          & 0.9707          & 0.8514                     & 0.8725          & 0.9686          & 0.8542           & 0.8862          & 0.9690          & 0.8657                  & 0.8793          & 0.9662          & 0.8571            & 0.8819          & 0.9686          \\
\texttt{T$^2$A} (Ours)                    & \textbf{0.8728}          & \textbf{0.8972} & \textbf{0.9723} & \textbf{0.8485}            & \textbf{0.8865} & \textbf{0.9699} & \textbf{0.8514}  & \textbf{0.8982} & \textbf{0.9740} & \textbf{0.8700}         & \textbf{0.8862} & \textbf{0.9689} & \textbf{0.8607}   & \textbf{0.8920} & \textbf{0.9713} \\
\textbf{Intensity level = 4}  &                          &                 &                 &                            &                 &                 &                  &                 &                 &                         &                 &                 &                   &                 &                 \\
Source                        & 0.7142                   & 0.8181          & 0.9502          & 0.6957                     & 0.6909          & 0.9011          & 0.8200           & 0.8019          & 0.9411          & 0.8057                  & 0.7876          & 0.9355          & 0.7589            & 0.7746          & 0.9319          \\
TENT                          & 0.8450                   & 0.8836          & 0.9677          & 0.7971                     & 0.7568          & 0.9316          & 0.8300           & 0.8372          & 0.9531          & \textbf{0.8428}         & 0.8433          & 0.9481          & 0.8287            & 0.8302          & 0.9501          \\
MEMO                          & 0.8085                   & 0.8306          & 0.9506          & \textbf{0.8085}            & 0.7502          & 0.9250          & 0.8271           & 0.8171          & 0.9482          & 0.8228                  & 0.8421          & 0.9535          & 0.8167            & 0.8100          & 0.9443          \\
EATA                          & 0.8442                   & 0.8737          & 0.9579          & 0.7957                     & 0.7658          & 0.9313          & 0.8285           & 0.8375          & 0.9532          & \textbf{0.8428}         & 0.8437          & 0.9482          & 0.8278            & 0.8301          & 0.9426          \\
CoTTA                         & 0.8400                   & 0.8352          & 0.9446          & 0.7585                     & 0.7158          & 0.9151          & \textbf{0.8357}  & 0.8323          & 0.9526          & 0.8257                  & 0.8357          & 0.9522          & 0.8150            & 0.8048          & 0.9411          \\
LAME                          & 0.8042                   & 0.7857          & 0.9296          & 0.8042                     & 0.6464          & 0.8706          & 0.8042           & 0.8010          & 0.9306          & 0.8042                  & 0.6581          & 0.8662          & 0.8042            & 0.7228          & 0.8993          \\
VIDA                          & \textbf{0.8528}          & 0.8581          & 0.9573          & 0.7700                     & 0.7173          & 0.9112          & 0.8257           & 0.8235          & 0.9456          & 0.8300                  & 0.8287          & 0.9482          & 0.8196            & 0.8069          & 0.9406          \\
COME                          & 0.8485                   & 0.8757          & 0.9657          & 0.7985                     & 0.7618          & 0.9342          & 0.8271           & 0.8279          & 0.9428          & 0.8400                  & 0.8514          & 0.9569          & 0.8285            & 0.8292          & 0.9424          \\
\texttt{T$^2$A} (Ours)                    & 0.8457                   & \textbf{0.8838} & \textbf{0.9678} & 0.8042                     & \textbf{0.7692} & \textbf{0.9426} & 0.8285           & \textbf{0.8376} & \textbf{0.9533} & 0.8414                  & \textbf{0.8536} & \textbf{0.9579} & \textbf{0.8300}   & \textbf{0.8360} & \textbf{0.9554} \\
\textbf{Intensity level = 5}  &                          &                 &                 &                            &                 &                 &                  &                 &                 &                         &                 &                 &                   &                 &                 \\
Source                        & 0.6614                   & 0.7780          & 0.9368          & 0.7085                     & 0.6616          & 0.8841          & 0.6800           & 0.8003          & 0.9438          & 0.8042                  & 0.6997          & 0.8995          & 0.7135            & 0.7349          & 0.9160          \\
TENT                          & ~0.8514                  & 0.8499          & 0.9475          & 0.7514                     & 0.7020          & 0.9051          & 0.7971           & 0.8119          & 0.9470          & 0.8171                  & 0.8067          & 0.9403          & 0.8042            & 0.7926          & 0.9349          \\
MEMO                          & 0.8157                   & 0.8205          & 0.9450          & 0.8057                     & 0.6936          & 0.9035          & 0.8185           & 0.7926          & 0.9417          & 0.8100                  & 0.7988          & 0.9399          & 0.8125            & 0.7764          & 0.9325          \\
EATA                          & 0.8500                   & 0.8500          & 0.9477          & 0.7514                     & 0.7021          & 0.9052          & 0.7929           & 0.8052          & 0.9372          & 0.8185                  & 0.8088          & 0.9433          & 0.8032            & 0.7915          & 0.9333          \\
CoTTA                         & 0.8114                   & 0.8019          & 0.9295          & 0.7285                     & 0.6772          & 0.8938          & 0.7742           & 0.7750          & 0.9340          & 0.8128                  & 0.7939          & 0.9385          & 0.7817            & 0.7620          & 0.9239          \\
LAME                          & 0.6700                   & 0.6607          & 0.8939          & 0.8042                     & 0.5566          & 0.8287          & 0.7557           & 0.7404          & 0.9172          & 0.8042                  & 0.5966          & 0.8438          & 0.7585            & 0.6385          & 0.8709          \\
VIDA                          & 0.8300                   & 0.8391          & 0.9484          & 0.7500                     & 0.6722          & 0.8860          & 0.800            & 0.7962          & 0.9395          & 0.8071                  & 0.7843          & 0.9310          & 0.7967            & 0.7730          & 0.9262          \\
COME                          & 0.8457                   & 0.8516          & 0.9540          & 0.7485                     & 0.6980          & 0.9066          & 0.8057           & 0.8040          & 0.9405          & 0.8157                  & 0.8098          & 0.9432          & 0.8039            & 0.7808          & 0.9260          \\
\texttt{T$^2$A} (Ours)                    & \textbf{0.8564}          & \textbf{0.8601} & \textbf{0.9577} & \textbf{0.7557}            & \textbf{0.7031} & \textbf{0.9054} & \textbf{0.7942}  & \textbf{0.8150} & \textbf{0.9469} & \textbf{0.8257}         & \textbf{0.8102} & \textbf{0.9440} & \textbf{0.8080}   & \textbf{0.7971} & \textbf{0.9385} 
\end{tblr}
\end{table*}

\definecolor{FrenchPass}{rgb}{0.741,0.878,0.996}
\begin{table*}[ht]
\centering
\fontsize{6pt}{6pt}\selectfont
\caption{Improvement of deepfake detectors to unknown postprocessing techniques from intensity level from 1 to 5.}
\label{tab:DF-post-process-appendix}
\begin{tblr}{
  width = \linewidth,
  colspec = {Q[80]Q[35]Q[35]Q[35]Q[35]Q[35]Q[35]Q[35]Q[35]Q[35]Q[35]Q[35]Q[35]Q[35]Q[35]Q[35]},
  cells = {c},
  row{5} = {FrenchPass},
  row{7} = {FrenchPass},
  row{9} = {FrenchPass},
  row{11} = {FrenchPass},
  row{14} = {FrenchPass},
  row{16} = {FrenchPass},
  row{18} = {FrenchPass},
  row{20} = {FrenchPass},
  row{23} = {FrenchPass},
  row{25} = {FrenchPass},
  row{27} = {FrenchPass},
  row{29} = {FrenchPass},
  row{32} = {FrenchPass},
  row{34} = {FrenchPass},
  row{36} = {FrenchPass},
  row{38} = {FrenchPass},
  row{41} = {FrenchPass},
  row{43} = {FrenchPass},
  row{45} = {FrenchPass},
  row{47} = {FrenchPass},
  cell{1}{1} = {r=2}{},
  cell{1}{2} = {c=3}{0.14\linewidth},
  cell{1}{5} = {c=3}{0.1\linewidth},
  cell{1}{8} = {c=3}{0.1\linewidth},
  cell{1}{11} = {c=3}{0.1\linewidth},
  cell{1}{14} = {c=3}{0.1\linewidth},
  cell{3}{1} = {c=16}{0.1\linewidth},
  cell{12}{1} = {c=16}{0.1\linewidth},
  cell{21}{1} = {c=16}{0.1\linewidth},
  cell{30}{1} = {c=16}{0.1\linewidth},
  cell{39}{1} = {c=16}{0.1\linewidth},
  vline{2,5,8,11,14} = {1-3, 4-11,13-20,22-29,31-38,40-47}{},
  hline{1,48} = {-}{0.08em},
  hline{2} = {2-16}{},
  hline{3-4,12-13,21-22,30-31,39-40} = {-}{},
  hline{3-4,12-13,21-22,30-31,39-40} = {2}{-}{},
}
\textbf{Method }              & \textbf{Color Saturation} &              &             & \textbf{Color Contrast} &              &             & \textbf{Gaussian Blur} &              &             & \textbf{Resize} &              &             & \textbf{Average} &              &             \\
                              & \textbf{ACC}              & \textbf{AUC} & \textbf{AP} & \textbf{ACC}            & \textbf{AUC} & \textbf{AP} & \textbf{ACC}           & \textbf{AUC} & \textbf{AP} & \textbf{ACC}    & \textbf{AUC} & \textbf{AP} & \textbf{ACC}     & \textbf{AUC} & \textbf{AP} \\
\textbf{Intensity level = 1~} &                           &              &             &                         &              &             &                        &              &             &                 &              &             &                  &              &             \\
CORE                          & 0.9000                    & 0.9441       & 0.9852      & 0.8957                  & 0.9444       & 0.9855      & 0.9042                 & 0.9423       & 0.9846      & 0.9028          & 0.9426       & 0.9847      & 0.9006           & 0.9434       & 0.9850      \\
CORE + \texttt{T$^2$A}                    & 0.8985                    & 0.9313       & 0.9784      & 0.9042                  & 0.9329       & 0.9794      & 0.8942                 & 0.9203       & 0.9729      & 0.8942          & 0.9214       & 0.9742      & 0.8977           & 0.9265       & 0.9762      \\
F3Net                         & 0.9028                    & 0.9629       & 0.9910      & 0.9114                  & 0.9634       & 0.9911      & 0.8971                 & 0.9570       & 0.9895      & 0.8900          & 0.9555       & 0.9891      & 0.9003           & 0.9597       & 0.9902      \\
F3Net + \texttt{T$^2$A}                   & 0.9071                    & 0.9594       & 0.9900      & 0.9057                  & 0.9607       & 0.9902      & 0.8942                 & 0.9554       & 0.9891      & 0.9042          & 0.9523       & 0.9881      & 0.9028           & 0.9570       & 0.9894      \\
RECCE                         & 0.8971                    & 0.9508       & 0.9871      & 0.9042                  & 0.9521       & 0.9875      & 0.8971                 & 0.9349       & 0.9824      & 0.8914          & 0.9357       & 0.9829      & 0.8975           & 0.9434       & 0.9850      \\
RECCE + \texttt{T$^2$A}                   & 0.8971                    & 0.9366       & 0.9825      & 0.8942                  & 0.9368       & 0.9827      & 0.8871                 & 0.9236       & 0.9779      & 0.8871          & 0.9242       & 0.9787      & 0.8914           & 0.9303       & 0.9805      \\
Effi. B4                      & 0.9100                    & 0.9615       & 0.9905      & 0.9100                  & 0.9607       & 0.9903      & 0.8971                 & 0.9401       & 0.9844      & 0.8957          & 0.9411       & 0.9847      & 0.9032           & 0.9509       & 0.9875      \\
Effi. B4 + \texttt{T$^2$A}                & 0.8871                    & 0.9393       & 0.9844      & 0.8914                  & 0.9396       & 0.984       & 0.8871                 & 0.9292       & 0.9813      & 0.8842          & 0.9287       & 0.9808      & 0.8875           & 0.9342       & 0.9826      \\
\textbf{Intensity level = 2}  &                           &              &             &                         &              &             &                        &              &             &                 &              &             &                  &              &             \\
CORE                          & 0.8814                    & 0.9272       & 0.9809      & 0.8242                  & 0.8861       & 0.9685      & 0.8542                 & 0.9176       & 0.9770      & 0.8628          & 0.9179       & 0.9780      & 0.8557           & 0.9122       & 0.9761      \\
CORE + \texttt{T$^2$A}                    & 0.8857                    & 0.9160       & 0.9749      & 0.8614                  & 0.8990       & 0.9705      & 0.8657                 & 0.8992       & 0.9679      & 0.8714          & 0.9001       & 0.9683      & 0.8711           & 0.9036       & 0.9704      \\
F3Net                         & 0.8985                    & 0.9359       & 0.9836      & 0.8557                  & 0.9205       & 0.9791      & 0.8685                 & 0.9293       & 0.9819      & 0.8857          & 0.9280       & 0.9805      & 0.8771           & 0.9284       & 0.9813      \\
F3Net + \texttt{T$^2$A}                   & 0.8928                    & 0.9512       & 0.9880      & 0.8757                  & 0.9246       & 0.9786      & 0.8700                 & 0.9193       & 0.9787      & 0.8871          & 0.9311       & 0.9885      & 0.8814           & 0.9316       & 0.9835      \\
RECCE                         & 0.8542                    & 0.9119       & 0.9756      & 0.8157                  & 0.8743       & 0.9640      & 0.8628                 & 0.8870       & 0.9670      & 0.8757          & 0.9073       & 0.9742      & 0.8521           & 0.8951       & 0.9702      \\
RECCE + \texttt{T$^2$A}                   & 0.8771                    & 0.9254       & 0.9796      & 0.8714                  & 0.9001       & 0.9717      & 0.8714                 & 0.8924       & 0.9675      & 0.8742          & 0.8981       & 0.9698      & 0.8735           & 0.9040       & 0.9722      \\
Effi. B4                      & 0.8857                    & 0.9422       & 0.9854      & 0.8171                  & 0.9202       & 0.9789      & 0.8785                 & 0.9009       & 0.9715      & 0.8728          & 0.9121       & 0.9759      & 0.8635           & 0.9189       & 0.9779      \\
Effi. B4 + \texttt{T$^2$A}                & 0.8885                    & 0.9190       & 0.9779      & 0.8871                  & 0.8985       & 0.9699      & 0.8728                 & 0.9001       & 0.9708      & 0.8628          & 0.9006       & 0.9729      & 0.8778           & 0.9046       & 0.9729      \\
\textbf{Intensity level = 3}  &                           &              &             &                         &              &             &                        &              &             &                 &              &             &                  &              &             \\
CORE                          & 0.8328                    & 0.8333       & 0.9514      & 0.8057                  & 0.8205       & 0.9401      & 0.8071                 & 0.8534       & 0.9538      & 0.8414          & 0.8714       & 0.9644      & 0.8218           & 0.8447       & 0.9524      \\
CORE + \texttt{T$^2$A}                    & 0.8514                    & 0.8639       & 0.9600      & 0.8600                  & 0.8716       & 0.9564      & 0.8457                 & 0.8576       & 0.9542      & 0.8485          & 0.8646       & 0.9595      & 0.8514           & 0.8644       & 0.9575      \\
F3Net                         & 0.8528                    & 0.8676       & 0.9627      & 0.7657                  & 0.8310       & 0.9509      & 0.8357                 & 0.8704       & 0.9644      & 0.8100          & 0.8800       & 0.9671      & 0.8161           & 0.8623       & 0.9613      \\
F3Net + \texttt{T$^2$A}                   & 0.8857                    & 0.9191       & 0.9771      & 0.8571                  & 0.8840       & 0.9641      & 0.8471                 & 0.8829       & 0.9660      & 0.8542          & 0.8882       & 0.9667      & 0.8610           & 0.8936       & 0.9685      \\
RECCE                         & 0.8242                    & 0.8014       & 0.9355      & 0.7928                  & 0.7976       & 0.9319      & 0.8171                 & 0.8201       & 0.9430      & 0.8300          & 0.8378       & 0.9505      & 0.8160           & 0.8142       & 0.9402      \\
RECCE + \texttt{T$^2$A}                   & 0.8385                    & 0.8498       & 0.9555      & 0.8457                  & 0.8622       & 0.9542      & 0.8371                 & 0.8496       & 0.9527      & 0.8442          & 0.8545       & 0.9524      & 0.8414           & 0.8540       & 0.9537      \\
Effi. B4                      & 0.8314                    & 0.8180       & 0.9394      & 0.6742                  & 0.8434       & 0.9539      & 0.8100                 & 0.8346       & 0.9463      & 0.8357          & 0.8414       & 0.9555      & 0.7878           & 0.8344       & 0.9488      \\
Effi. B4 + \texttt{T$^2$A}                & 0.8500                    & 0.8352       & 0.9460      & 0.8542                  & 0.8588       & 0.9511      & 0.8485                 & 0.8583       & 0.9556      & 0.8314          & 0.8492       & 0.9529      & 0.8460           & 0.8504       & 0.9514      \\
\textbf{Intensity level = 4}  &                           &              &             &                         &              &             &                        &              &             &                 &              &             &                  &              &             \\
CORE                          & 0.7542                    & 0.6845       & 0.8978      & 0.7871                  & 0.7571       & 0.9008      & 0.8000                 & 0.7604       & 0.9175      & 0.8071          & 0.8188       & 0.9438      & 0.7871           & 0.7552       & 0.9150      \\
CORE + \texttt{T$^2$A}                    & 0.7942                    & 0.7367       & 0.9136      & 0.8442                  & 0.8467       & 0.9530      & 0.8385                 & 0.8256       & 0.9434      & 0.8142          & 0.8043       & 0.9374      & 0.8228           & 0.8033       & 0.9369      \\
F3Net                         & 0.8057                    & 0.7018       & 0.8991      & 0.7428                  & 0.7486       & 0.9152      & 0.8057                 & 0.7542       & 0.9232      & 0.7671          & 0.8203       & 0.9442      & 0.7803           & 0.7562       & 0.9204      \\
F3Net + \texttt{T$^2$A}                   & 0.8185                    & 0.7970       & 0.9366      & 0.8414                  & 0.8448       & 0.9457      & 0.8185                 & 0.8501       & 0.9578      & 0.8285          & 0.8261       & 0.9453      & 0.8267           & 0.8295       & 0.9464      \\
RECCE                         & 0.7957                    & 0.6581       & 0.8743      & 0.7657                  & 0.7559       & 0.9135      & 0.8042                 & 0.7453       & 0.9125      & 0.7671          & 0.7782       & 0.9286      & 0.7832           & 0.7344       & 0.9072      \\
RECCE + \texttt{T$^2$A}                   & 0.7728                    & 0.7158       & 0.9000      & 0.8242                  & 0.8358       & 0.9458      & 0.8200                 & 0.8138       & 0.9403      & 0.8128          & 0.7884       & 0.9316      & 0.8075           & 0.7885       & 0.9295      \\
Effi. B4                      & 0.8157                    & 0.6537       & 0.8700      & 0.5628                  & 0.7847       & 0.9320      & 0.8028                 & 0.7005       & 0.8951      & 0.7928          & 0.7899       & 0.9313      & 0.7435           & 0.7322       & 0.9071      \\
Effi. B4 + \texttt{T$^2$A}                & 0.7685                    & 0.7174       & 0.9087      & 0.8271                  & 0.8281       & 0.9393      & 0.8228                 & 0.7966       & 0.9342      & 0.8128          & 0.7923       & 0.9373      & 0.8078           & 0.7836       & 0.9299      \\
\textbf{Intensity level = 5}  &                           &              &             &                         &              &             &                        &              &             &                 &              &             &                  &              &             \\
CORE                          & 0.7500                    & 0.6444       & 0.8823      & 0.7642                  & 0.7142       & 0.8794      & 0.8014                 & 0.6590       & 0.8717      & 0.7657          & 0.7633       & 0.9281      & 0.7703           & 0.6952       & 0.8904      \\
CORE + \texttt{T$^2$A}                    & 0.7771                    & 0.7008       & 0.8968      & 0.8342                  & 0.8238       & 0.9444      & 0.8042                 & 0.7784       & 0.9312      & 0.7914          & 0.7602       & 0.9178      & 0.8017           & 0.7658       & 0.9226      \\
F3Net                         & 0.8114                    & 0.6300       & 0.8701      & 0.7428                  & 0.6893       & 0.8826      & 0.8000                 & 0.6500       & 0.8809      & 0.7185          & 0.7778       & 0.9351      & 0.7682           & 0.6868       & 0.8922      \\
F3Net + \texttt{T$^2$A}                   & 0.8014                    & 0.7421       & 0.9078      & 0.8228                  & 0.8259       & 0.9419      & 0.7928                 & 0.7906       & 0.9380      & 0.7957          & 0.7811       & 0.9346      & 0.8032           & 0.7849       & 0.9306      \\
RECCE                         & 0.8028                    & 0.6351       & 0.8688      & 0.7614                  & 0.7144       & 0.8959      & 0.7985                 & 0.6808       & 0.8823      & 0.7042          & 0.7103       & 0.9055      & 0.7667           & 0.6852       & 0.8881      \\
RECCE + \texttt{T$^2$A}                   & 0.7600                    & 0.6882       & 0.8853      & 0.8157                  & 0.8143       & 0.9391      & 0.7928                 & 0.7651       & 0.9237      & 0.7857          & 0.7479       & 0.9151      & 0.7886           & 0.7539       & 0.9158      \\
Effi. B4                      & 0.8028                    & 0.6109       & 0.8458      & 0.5257                  & 0.7232       & 0.9102      & 0.8014                 & 0.5882       & 0.8454      & 0.7600          & 0.7447       & 0.9159      & 0.7225           & 0.6668       & 0.8793      \\
Effi. B4 + \texttt{T$^2$A}                & 0.7371                    & 0.6451       & 0.8691      & 0.8028                  & 0.7952       & 0.9259      & 0.7957                 & 0.7515       & 0.9160      & 0.7685          & 0.7588       & 0.9188      & 0.776            & 0.7377       & 0.9075      
\end{tblr}
\end{table*}

\end{document}